\documentclass[10pt]{article} % For LaTeX2e
\usepackage[preprint]{tmlr}

\definecolor{citecolor}{HTML}{0071bc}
\usepackage[colorlinks=true,allcolors=citecolor]{hyperref}
\usepackage{url}
\usepackage{amsmath}
\newcommand{\myMethod}{GRAPES}

\DeclareMathOperator*{\topk}{\operatorname{top-\!}k}
\usepackage{colortbl}
\usepackage{tabularx}
\usepackage{microtype}
\usepackage{enumitem}
\usepackage{arydshln}
\usepackage{wrapfig}
\usepackage{amsthm}
\usepackage{amssymb}
\usepackage{caption}
\usepackage{subcaption}
\usepackage[pdftex]{graphicx}
\usepackage{algorithm}%
\usepackage{algorithmicx}%
\usepackage[noend]{algpseudocode}%

\usepackage{booktabs}						% professional-quality tables
\usepackage{multirow}						% tabular cells spanning multiple rows
\usepackage{amsfonts}						% blackboard math symbols
\usepackage{graphicx}						% figures
\usepackage{duckuments}						% sample images

\title{GRAPES: Learning to Sample Graphs for \\
Scalable Graph Neural Networks}

\author{\name Taraneh Younesian \email t.younesian@vu.nl \\
      \addr Vrije Universiteit Amsterdam
      \AND
      \name Daniel Daza \email d.f.dazacruz@amsterdamumc.nl \\
      \addr Amsterdam UMC
      \AND
      \name Emile van Krieken \email Emile.van.Krieken@ed.ac.uk \\
      \addr University of Edinburgh\\
      \AND
      \name Thiviyan Thanapalasingam \email t.singam@uva.nl\\ \addr University of Amsterdam 
      \AND
      \name Peter Bloem \email p.bloem@vu.nl \\ \addr Vrije Universiteit Amsterdam
      }

\newtheorem{theorem}{Theorem}
\newtheorem{definition}{Definition}

\def\month{05}  % Insert correct month for camera-ready version
\def\year{2025} % Insert correct year for camera-ready version
\def\openreview{\url{https://openreview.net/forum?id=QI0l842vSq}} % Insert correct link to OpenReview for camera-ready version

\begin{document}

\maketitle

\begin{abstract}
Graph neural networks (GNNs) learn to represent nodes by aggregating information from their neighbors. As GNNs increase in depth, their receptive field grows exponentially, leading to high memory costs. 
Several works in the literature proposed to address this shortcoming by sampling subgraphs or by using historical embeddings. These methods have mostly focused on benchmarks of single-label node classification on \emph{homophilous graphs}, where neighboring nodes often share the same label.
However, most of these methods rely on static heuristics that may not generalize across different graphs or tasks. We argue that the sampling method should be \emph{adaptive}, adjusting to the complex structural properties of each graph.
To this end, we introduce \myMethod{}, an adaptive sampling method that learns to identify the set of nodes crucial for training a GNN.
\myMethod{} trains a second GNN to predict node sampling probabilities by optimizing the downstream task objective.
We evaluate \myMethod{} on various node classification benchmarks involving homophilous as well as heterophilous graphs. We demonstrate \myMethod{}' effectiveness in accuracy and scalability, particularly in multi-label heterophilous graphs. Additionally, \myMethod{}
uses orders of magnitude less GPU memory than a strong baseline based on historical
embeddings.
Unlike other sampling methods, \myMethod{} maintains high accuracy even with smaller sample sizes and, therefore, can scale to massive graphs. Our implementation is publicly available online.\footnote{Available at \url{https://github.com/dfdazac/grapes}.}
\end{abstract}

\section{Introduction}\label{sec1}
% In many applications, data is represented with graph structures, such as in recommender systems, social networks, and the chemical and medical domains \citep{nettleton2013data, wu2022graph, li2022graph}. In these domains, graph neural networks (GNNs) are a powerful tool for representation learning on graphs \citep{kipf2016semi, velickovic2017graph, yun2019graph}. 
    
Despite the broad range of applications of GNNs \citep{nettleton2013data, wu2022graph, li2022graph, kipf2016semi, velickovic2017graph, yun2019graph}, scalability remains a significant challenge \citep{serafini2021scalable}. Unlike traditional machine learning problems, where data is assumed to be i.i.d., the graph structure introduces dependencies between a node and its neighborhood. This complicates partitioning data into mini-batches. Additionally, the number of nodes that a GNN needs to process increases exponentially with the number of layers \citep{ma2025acceleration}. 

Most current graph sampling methods use a fixed heuristic to compute node inclusion probabilities: The sampling process is independent of the node features, graph structure, or the GNN task performance \citep{fastGCN,LADIES, graphsaint}. For simple graphs like \emph{homophilous graphs} \citep{zhu2020beyond,zheng2022graph,zhao2023multi} where there is a strong correlation between labels of a node and its neighbors, a fixed heuristic typically suffices. 
% However, such methods cannot adapt to more complicated graph structures, where the neighborhood contains a diverse set of nodes, while only some neighbors provide important information about the class of the target node.
However, these methods fall short when applied to graphs in which only a few of the nodes in the neighborhood provide relevant information for the task at hand.
Consequently, we study \emph{adaptive sampling}, where the sampling method adapts dynamically to the task by learning which nodes should be included. 
Our adaptive sampling technique learns to sample by directly minimizing the classification loss of the downstream GNN. This differs from most existing graph sampling methods, which aim to approximate the full-batch GNN~\citep{fastGCN, LADIES, graphsaint, AS-GCN}, opting for an indirect approach to achieve high accuracy.

\begin{figure}[t]%
    \centering
    \includegraphics[width=\textwidth, clip]{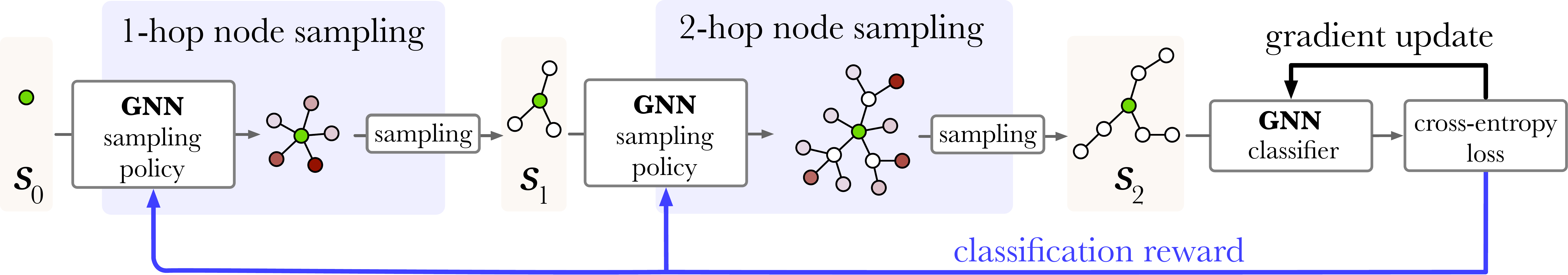}
    \caption{Overview of \myMethod{}. First, \myMethod{} processes a target node (green) by computing node inclusion probabilities on its 1-hop neighbors (shown by node color shade) with a sampling GNN. Given these probabilities, \myMethod{} samples $k$ nodes. Then, \myMethod{} repeats this process over nodes in the 2-hop neighborhood. We pass the sampled subgraph to the classifier GNN for target node classification. Finally, \myMethod{} uses the classification loss to update the classifier GNN and to reward the sampler GNN.}
    \label{fig:grapes-architecture}
\end{figure}

We introduce a straightforward adaptive sampling method called \textbf{Gr}aph \textbf{A}da\textbf{p}tiv\textbf{e} \textbf{S}ampling (\myMethod{}). As illustrated in Figure~\ref{fig:grapes-architecture}, \myMethod{} samples a subgraph around the target nodes in a series of steps. At each step, a \emph{sampling policy} GNN computes inclusion probabilities for the nodes neighboring the current subgraph, which the sampler uses to select a subset. Once the sampling is complete, the resulting subgraph is passed to a second \emph{classifier} GNN for classification. The classification loss is then backpropagated to train both GNNs.

To enable backpropagation through the sampling step, we need a gradient estimator. We compare a reinforcement learning (RL) approach and a GFlowNet (GFN) approach \citep{bengio2021flow}. Both approaches allow us to consider the sampling process and the GNN computation together and to train them concurrently.
This allows the sampler to adapt based on factors like the node features, the graph structure, the sample size, and other contextual features. 

\begin{figure}[t]
    \centering
    \hspace*{\fill}
    \begin{subfigure}[b]{0.35\textwidth}
        \centering
        \includegraphics[width=\textwidth]{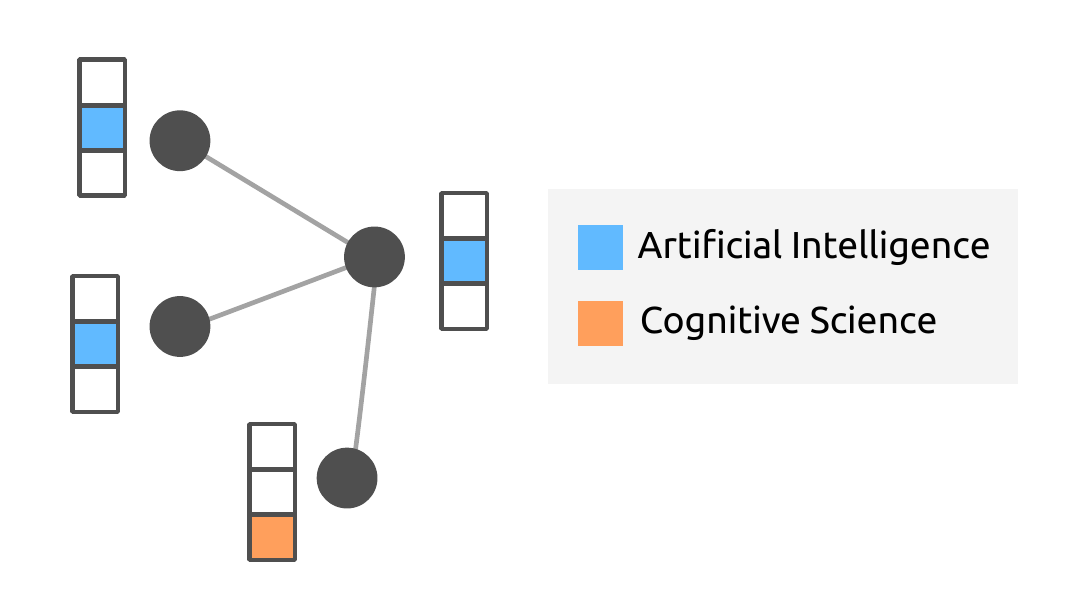}
        \caption{}
        \label{subfig:mchomo}
    \end{subfigure}
    \hspace*{\fill}
    \begin{subfigure}[b]{0.35\textwidth}
        \centering
        \includegraphics[width=\textwidth]{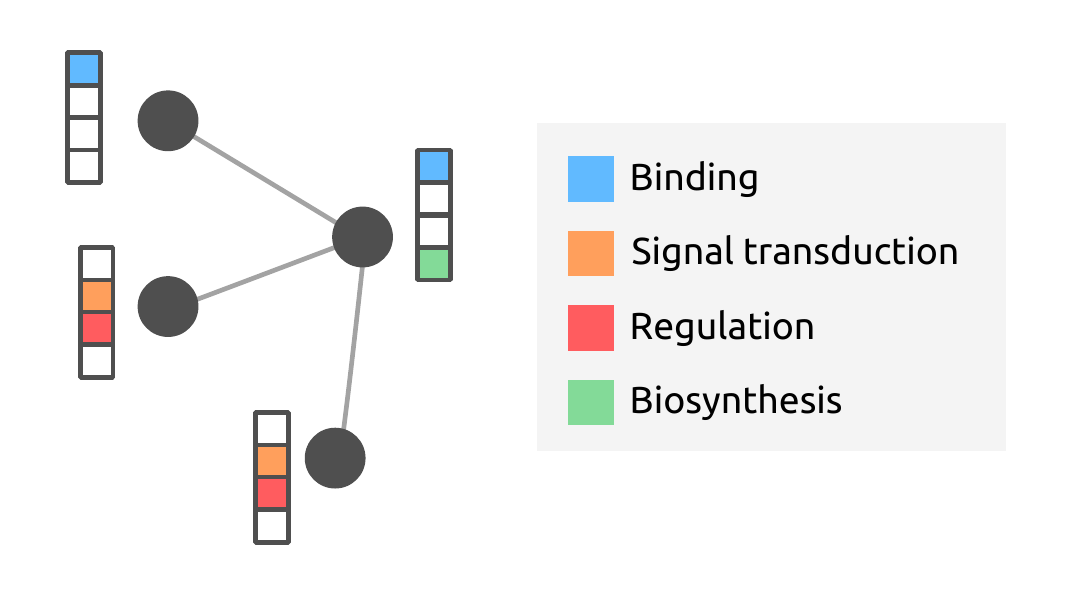}
        \caption{}
        \label{subfig:mlhetero}
    \end{subfigure}
    \hspace*{\fill}
    \caption{The downstream task and the structural properties of the graph affect the strategy used for sampling. Fig.~\ref{subfig:mchomo} shows a citation network where articles citing each other are likely to belong to the same category. Fig.~\ref{subfig:mlhetero} shows a graph of interacting proteins, each with different biological functions.}
    \label{fig:mainfig}
\end{figure}

%Scaling GNNs to large graphs through sampling presents unique challenges, determined by both the complexity of the downstream task and the structural properties of the graph. 
In multi-class classification-- where a node can be assigned only one of multiple classes-- on homophilous graphs, even a simple strategy like random sampling could be sufficient, since the neighborhood of a node contains nodes with similar labels (see Fig.~\ref{subfig:mchomo}). In the case of multi-label classification, on the other hand, the number of possible labels assigned to a node is $\vert2^{\mathcal{Y}}\vert$, where $\mathcal{Y}$ is the unique set of labels. Furthermore, in heterophilous graphs, the relationship between the neighborhood of a node and its label is more complex (see Fig.~\ref{subfig:mlhetero}). We argue that in such cases, a sampling policy that directly adapts to the properties of the graph and the downstream task is crucial. 

Therefore, in addition to the datasets commonly used in the literature, we evaluate \myMethod{} on several heterophilous and/or multi-label graphs and demonstrate its effectiveness on these complicated graphs. Moreover, we spend significant effort on the evaluation protocol for fairly testing graph sampling methods. In particular, we ensure that all sampling methods are evaluated under the same conditions and on the same GNN architecture to eliminate any confounding factors and to ensure that any changes in performance can only be attributed to the sampling method. To the best of our knowledge, we are the first to perform such a rigorous comparison on twelve varied datasets.

Additionally, we provide a theoretical analysis indicating the effectiveness of adaptive sampling on a specific category of heterophilous graphs, where only a subset of neighbors contains relevant information for an accurate node classification. %, and we show that these graphs are heterophilous. 

We evaluate \myMethod{} on several node classification tasks and find that  
% \begin{enumerate}
% \item \textbf{Competitive Performance.} \myMethod{} is competitive to other sampling methods in accuracy across datasets. On heterophilous and multi-label datasets \myMethod{} provides state-of-the-art performance. 
% %\item \textbf{Homophilous Graphs.} For some benchmarks, mostly those based on homophilous graphs, simple baselines like random sampling perform as well as any state-of-the-art method, including \myMethod{}.
% \item \textbf{Heterophilous Graphs.} \myMethod{} outperforms all the baselines for all the heterophilous -- and multi-label-- benchmarks, indicating the necessity of using \myMethod{} for complex graphs.
% \item \textbf{Memory Efficiency.} \quad \myMethod{} is competitive with a state-of-the-art non-sampling GNN scaling method, while using up to an order of magnitude less GPU memory.
% \end{enumerate}

% \paragraph{Alternative suggestion} We evaluate \myMethod{} on several node classification benchmarks and find the following:
\begin{enumerate}
\item \textbf{Performance.} \myMethod{} achieves state-of-the-art performance in multi-label classification benchmarks over heterophilous graphs, while performing competitively on multi-class classification on homophilous graphs.
\item \textbf{Memory Efficiency.} \myMethod{} uses orders of magnitude less GPU memory when compared with a strong baseline that relies on historical embeddings for scaling GNN training.
\item \textbf{Robustness.} In comparison with other methods, \myMethod{} maintains good performance under exponential reductions in sample size.
\end{enumerate}

\section{Related Work}
\textbf{Fixed Sampling Policy:} In this category, the sampling policy is independent of the training of the GNN and is based on a fixed heuristic that does not involve any training. Given a set of target nodes, \emph{i.e.} the nodes that are to be classified, node-wise sampling methods sample a given number of nodes for each target node. GraphSage \citep{hamilton2017inductive} is a node-wise sampling method that randomly samples nodes. However, node-wise sampling can result in nodes being sampled multiple times redundantly because they can be the neighbors of several nodes \citep{LADIES}. 
% A more efficient approach is sampling nodes per layer, called layer-wise sampling. FastGCN \citep{fastGCN} and LADIES \citep{LADIES} are examples of this approach, and so is \myMethod{}. They aim to minimize variance by sampling nodes with probabilities proportional to their degree.
A more efficient approach is layer-wise sampling, for example, FastGCN \citep{fastGCN} and LADIES \citep{LADIES}. They aim to minimize variance by sampling nodes in each layer with probabilities proportional to their degree. MVS-GNN~\citep{cong2020minimal} proposes a fixed sampling policy with the aim of minimizing sampling variance by decoupling it into two components. The first component, embedding approximation variance, arises from neighbor sampling and is mitigated through the use of historical embeddings. The second component, stochastic gradient variance, results from mini-batching and is addressed by incorporating the norm of the node gradients into the sampling process. Moreover, some techniques focus on sampling subgraphs in each mini-batch, like GraphSAINT~\citep{graphsaint} and ClusterGCN~\citep{ClusterGCN}. While these techniques effectively scale GNNs to larger graphs, they do not adapt to the sampling policy based on the GNN's performance on the task. In the graph signal processing community, the authors in \citep{geng2023pyramid} propose a node sampling technique based on \citep{anis2016efficient} to sample nodes for a unique and stable graph signal reconstruction. However, this method is only applied to small graphs.

\textbf{Learnable Sampling Policy:} A few methods learn the probability of including a node based on feedback from the GNN. AS-GCN \citep{AS-GCN} is a method that learns a linear function that estimates the node probabilities layer-wise. Similarly, PASS \citep{PASS} learns a mixture of a random distribution and a learned policy with RL. Like GRAPES, it is task-adaptive; however, its sampling probability uses a bilinear similarity measure between neighboring node features, meaning the graph structure is not considered. FairSample \cite{cong2023fairsample} is a learnable sampling policy that borrows the PASS training mechanism that utilizes reinforcement learning to combine classification loss and a fairness loss to balance accuracy and fairness towards the underrepresented nodes.
GNN-BS \citep{gnn-bs} formulates the node-wise sampling problem as a bandit problem and updates the sampling policy according to a reward function that reduces the sampling variance. SubMix \citep{abu2023submix} proposes a mixture distribution of sampling heuristics with learnable mixture weights.
DSKReG \citep{wang2021dskreg} learns the relevance of items in a user-item knowledge graph by jointly optimizing the sampling strategy and the recommender model. The majority of these methods focus on variance reduction and fail to consider the classification loss, unlike \myMethod{}. We argue that adaptivity to the classification loss allows for sampling the influential nodes depending on the task and results in better performance.

\textbf{Other Scalable Methods:} Authors of \citep{you2022early, zhang2024lottery} use graph lottery tickets to eliminate unnecessary edges, while DSpar \citep{liu2023dspar} uses a simple degree-based heuristic to sparsify the graph. \citet{ruiz2023transferability} proposes a method to transfer the weights of a GNN trained on a mid-sized graph to larger graphs, given the graphon similarity between the graphs.
Another group of papers uses historical embeddings of the nodes when updating the target nodes' embeddings \citep{vr-gcn, fey2021gnnautoscale, yu2022graphfm, shi2023lmc}. %VR-GCN \citep{vr-gcn} uses historical activations with the aim of variance reduction. 
GAS \citep{fey2021gnnautoscale} approximates the embeddings of the 1-hop neighbors using the historical embeddings of those nodes learned in the previous training iterations. These methods reduce the GPU memory usage by training in mini-batches and learning from the 1-hop neighbors with the historical embeddings saved in CPU memory. Unlike \myMethod{}, they process the whole graph.

\section{Background: GNN Training and Sampling}
We first provide the necessary background about GNNs and graph sampling. Although our method is independent of the choice of GNN architecture, we limit our discussion to the GCN architecture for simplicity \citep{kipf2016semi}. 

Let $\mathcal{G}=(\mathcal{V,E})$ be an undirected graph with a list of $N$ nodes $\mathcal{V}=\{1, \ldots, N\}$ and a set of edges $\mathcal{E}$. The adjacency matrix $A\in \{0, 1\}^{N\times N}$ indicates a connection between a pair of nodes. Next, let $\hat{A}=\tilde{D}^{-1/2}\tilde{A}\tilde{D}^ {-1/2}$, where $\tilde{A}=A+I$ and where $\tilde{D}$ is the degree matrix of $\tilde{A}$. Let $X \in \mathbb{R}^{N \times f}$ be the node embeddings and let $Y$ be the labels for the \emph{target nodes}
%$\{(X^t_i, Y_i)\}_{i=1}^{N^t}$ indicate the set of features and labels for the target nodes
$\mathcal{V}^t \subset \mathcal{V}$, where $\mathcal{V}^ t$ indexes the nodes with a label. 
%In node classification, we want to learn to predict the labels of the target nodes. 

We consider a GCN with $L$ layers. The output of the $l$-th layer of the GCN is $H^{(l)} =  \sigma (\tilde{A}H^{(l-1)}W^{(l)})$, where $W^{(l)}$ is the weight matrix of GCN layer $l$ and $\sigma$ is a non-linear activation function. For a node $i \in \mathcal{V}$, this corresponds to the update
\begin{equation}
    h^{(l)}_{i} = \sigma \left( \sum_{j \in \mathcal{N}(i) \cup \{i\}}\hat{A}_{ij}h^{(l-1)}_{j}W^{(l)} \right),
\end{equation}
where $\mathcal{{N}}(i): \mathcal{V}\rightarrow 2^ {\mathcal{V}}$ is the set of $i$'s neighbors excluding $i$.

As the number of layers increases, the computation of the embedding of the node $i$ involves neighbors from further hops. 
% The more layers are added to the network, the further hop neighbors of each node will be included in the training. 
As a result, the neighborhood size grows rapidly with the number of layers.  
We study how to sample the graph to overcome this growth. We focus on layer-wise sampling, a common type of graph sampling approach \citep{fastGCN,LADIES,AS-GCN}.
% To overcome the neighborhood explosion problem, in this paper, we will focus on layer-wise sampling, which is an effective graph sampling approach \citep{fastGCN, LADIES, AS-GCN}. 
First, we divide the target nodes into mini-batches of size $b$. Then, in each layer, we sample $k$ nodes $\mathcal{V}^{(l)}$ among the neighbors of the nodes in the previous layer using the sampling policy $q$. To make this precise, we will need some additional notation. The computation of the output of layer $l$ is then:
\begin{equation}
\begin{aligned}
    \label{eq:approx}
    \text{for all $i \in K^{(l)}$,}&\quad h'^{(l)}_{i} = \sigma \left ( \sum_{j \in K^{(l-1)}} \hat{A}^{(l)}_{ij}h'^{(l-1)}_{j}W^{(l)}  \right ) \\ 
    \text{with}&\quad\mathcal{V}^{(l-1)}, \mathcal{V}^{(l)} \sim q(\mathcal{V}^{(l-1)}, \mathcal{V}^{(l)}|k),
\end{aligned}
\end{equation}
where 
\begin{enumerate}[noitemsep]
    \item $K^{(0)}=\mathcal{V}^{(0)}$ is the set of target nodes in the current mini-batch, where $\mathcal{V}^{(0)} \subset \mathcal{V}^t$;
    \item $K^{(l)}=\mathcal{V}^{(l)}\cup \mathcal{V}^{(0)}$ for all $l\in 1, \ldots, L$ adds the batch nodes $\mathcal{V}^{(0)}$ to the sampled nodes $\mathcal{V}^{(l)}$ to ensure self-loops between the batch nodes; 
    \item $\mathcal{V}^{(l)}\subseteq\mathcal{N}(K^{(l-1)})$ are the nodes sampled in layer $l$ among the neighbors $\mathcal{N}(K^{(l-1)})$ of the nodes in $K^{(l-1)}$. Note that $\mathcal{V}^{(l)}$ cannot contain nodes in $K^{(l-1)}$; 
    \item $\hat{A}^{(l)}=D^ {(l)^ {-1/2}}A^ {(l)}D^ {(l)^{-1/2}}$ is computed from the adjacency matrix $A^{(l)}$ containing the edges between $K^{(l)}$ and $K^{(l-1)}$, and the corresponding degree matrix $D^{(l)}$.\footnotemark\ To be precise, the entries $A^{(l)}_{ij}$ are 1 if and only if $i \in K^{(l)}$, $j \in K^{(l-1)}$ and if there is an edge $(i, j)$ in the original graph, that is, $\tilde{A}_{ij}=1$;
    \item $q$ is a sampling policy that samples the 
 $k$ nodes $\mathcal{V}^{(l)}$. 
\end{enumerate}

\footnotetext{Unlike the full-batch GCN, the adjacency matrix varies across layers when sampling because each layer involves a different set of nodes. Note also that, unlike the full-batch setting, message passing is \emph{asymmetric}: node $i$ may be updated from node $j$ but not vice versa.}

Existing layer-wise sampling methods, such as LADIES or FastGCN, use a fixed heuristic to determine $q$, for instance, by computing node probabilities proportional to the node degrees. However, an adaptive method \emph{learns} the distribution $q$ instead.

\section{Graph Adaptive Neighbor Sampling (\myMethod)}
We introduce Graph Adaptive Sampling (\myMethod{}). \myMethod{} is a layer-wise and layer-dependent adaptive sampling method that learns a sampling policy that minimizes the training objective conditioned on the input graph. In each layer $l$, we sample a subset $\mathcal{V}^{(l)}$ that is much smaller than the neighborhood of the previous layer. That is, $|\mathcal{V}^{(l)}|=k\ll |\mathcal{N}(K^{(l-1)})|$, where again $K^{(l-1)}=\mathcal{V}^{(0)}\cup \mathcal{V}^{(l-1)}$ adds the batch nodes to the sampled nodes of the previous layer. We use a second GNN to compute the inclusion probability for each node in $\mathcal{N}(K^{(l-1)})$. In the remainder of this section, we describe our sampling policy $q$ and training methods for the sampling policy. Furthermore, Algorithm \ref{algo1} shows one epoch of \myMethod{} in pseudocode.

\subsection{Sampling policy}
\label{sec:sampling-policy}
Next, we specify the sampling policy GNN  $\operatorname{GCN}_{\operatorname{S}}(K^{(l-1)})$, which computes inclusion probabilities on the subgraph created from the nodes in $K^{(l-1)}$ and their neighbors $\mathcal{N}^{(l-1)}$. The sampling policy $q$ decomposes as $q(\mathcal{V}^{(1)}, \ldots, \mathcal{V}^{(L)}|\mathcal{V}^{(0)})=\prod_{l=1}^L q(\mathcal{V}^{(l)}|\mathcal{V}^{(0)}, \ldots, \mathcal{V}^{(l-1)})$. We compute each factor $q(\mathcal{V}^{(l)}|\mathcal{V}^{(0)}, \ldots, \mathcal{V}^{(l-1)})$ as a product of Bernoulli inclusion probabilities, given by $\operatorname{GCN}_{\operatorname{S}}$, for each node $i$ in the neighborhood $\mathcal{N}(K^{(l-1)})$:
\begin{equation}
    q(\mathcal{V}^{(l)}|\mathcal{V}^{(0)}, \ldots, \mathcal{V}^{(l-1)})=\prod_{i\in \mathcal{N}(K^{(l-1)})} \operatorname{Bern}(i\in \mathcal{V}^{(l)}|p_i), \quad p_i=\operatorname{GCN}_{\operatorname{S}}(K^{(l-1)})_i.
\end{equation}
In addition to the regular embeddings $X$, the sampler GNN also has access to a one-hot vector of length $L+1$ that records the value $l+1$ for nodes sampled in layer $l$ (with 1 recorded for the target nodes). This allows the sampler to differentiate between nodes sampled in different layers.

\subsection{Sampling exactly k nodes}
\label{sampling and off policy}
We have a clear constraint on the number of nodes we want to include in training: In Equation \ref{eq:approx}, we sample exactly $k$ nodes without replacement. However, our sampling policy $q(\mathcal{V}^{(l)}|\mathcal{V}^{(0)}, \ldots, \mathcal{V}^{(l-1)})$ consists of many independent Bernoulli distributions, and it is highly unlikely that we sample exactly $k$ nodes from this distribution. 

Instead, we use the Gumbel-Top-$k$ trick \citep{vieira2014gumbel, huijben2022review-gumbel}, which selects a set of exactly $k$ nodes $\mathcal{V}^{(l)}$ by perturbing the log probabilities randomly and taking the top-$k$ among those:
\begin{equation}
    \mathcal{V}^{(l)} = \topk_{i\,\in\,\mathcal{N}(K^{(l-1)})} \log p_i+\epsilon_i, \quad \epsilon_i \sim \operatorname{Gumbel}(0,1)
    \label{gumbel}
\end{equation}
This guarantees a sample from $q(\mathcal{V}^{(l)}|\mathcal{V}^{(0)},\ldots, \mathcal{V}^{(l-1)}, k)$ that conditions on the number of nodes sampled.

While this results in a tractable and adaptive sampling procedure of exactly $k$ nodes for the classifier GNN, we have not yet given a method for learning the sampling policy. Unfortunately, a sampling operation, or in this case, the top-$k$ operation, provides no functional gradient \citep{mohamedMonteCarloGradient2020}. We will resort to simple methods from the reinforcement learning and GFlowNet literature to still be able to train the sampling policy, which we will explain next.

The choice of a Bernoulli distribution for sampling node neighborhoods is motivated by the idea that when conditioned on a target node, we can sample a node independent of other nodes in the neighborhood to maintain efficiency. Other assumptions could be more elaborate, such as sampling jointly pairs of nodes at a time, but this would in turn increase the computational cost of computing probabilities and sampling. On the other hand, the Gumbel top-k trick allows us to sample exactly $k$ elements without replacement, and while there are other applicable methods such as reservoir sampling, they are mathematically equivalent~\citep{huijben2022review-gumbel}.

\subsection{Training the Sampling Policy}
We train the sampling policy to minimize the classification loss $\mathcal{L}_{\operatorname{C}}$ of the classifier GNN\footnotemark. We use two methods to train the sampling policy $\operatorname{GCN}_{\operatorname{S}}$: a reinforcement learning (RL) method and a GFlowNet (GFN) method. We also experimented with the straight-through estimator~\citep{bengio2013st} for learning in \myMethod{}, which we observed not to perform well due to increased computation and high bias in the estimated gradients (see Appendix \ref{sec:app-sthrough}).

\footnotetext{\myMethod{} can be extended to other tasks, but we focus on node classification in the current work.}

\paragraph{REINFORCE (\myMethod{}-RL)} In the REINFORCE-based method, we use a simple REINFORCE estimator \citep{williamsSimpleStatisticalGradientfollowing1992} to compute an unbiased gradient of the classification loss. 
\begin{equation}
    \label{eq:rl}
    \mathcal{L}_{\operatorname{RL}}(X,Y, \mathcal{V}^ {(0)}) = \mathcal{L}_{\operatorname{C}}(X,Y, K^ {(0)},  \ldots, K^ {(L)}) \log q(\mathcal{V}^ {(1)}, \ldots, \mathcal{V}^ {(L)}|\mathcal{V}^ {(0)}),
\end{equation}
where we sample $\mathcal{V}^ {(1)}, \ldots, \mathcal{V}^ {(L)} \sim q(\mathcal{V}^ {(1)}, \ldots, \mathcal{V}^ {(L)}|\mathcal{V}^ {(0)}, k)$. Taking the derivative with respect to the parameters of $q$ results in the standard REINFORCE estimator. Note that this is an off-policy estimator since we sample from $q$ conditioned on the number of samples $k$ with the Gumbel-Top-$k$ trick, but compute gradients with respect to the distribution unconditioned on $k$. This is because computing likelihoods conditioned on $k$, although possible \citep{ahmed2023simple}, is computationally expensive. We discuss this issue in more detail in Appendix \ref{appendix:off-policy}. 

\paragraph{GFlowNets (\myMethod{}-GFN)} The second method uses the Trajectory Balance loss~\citep{malkin2022trajectorybalance} from the GFlowNet literature~\citep{bengio2021gflow-fundation,bengio2021flow}, which is known to perform well in off-policy settings \citep{malkin2022gflownets-off}. 
\begin{equation}
    \label{eq:gfn}
    \begin{aligned}
    \mathcal{L}_{\operatorname{GFN}}(X,Y, \mathcal{V}^ {(0)}) = \Big(&\log Z(\mathcal{V}^{(0)}) + \log q(\mathcal{V}^ {(1)}, \ldots, \mathcal{V}^ {(L)}|\mathcal{V}^ {(0)}) \\
    &+ \alpha\cdot \mathcal{L}_{\operatorname{C}}(X,Y, K^ {(0)},  \ldots, K^ {(L)}) \Big)^2,
    \end{aligned}
\end{equation}
where $\log Z(\mathcal{V}^{(0)})$ is a small GCN that predicts a scalar from the target nodes, and $\alpha$ is a tunable reward scaling hyperparameter. For a detailed derivation, see Appendix \ref{sec:gflownet-design}. GFlowNets minimize an objective that ensures sampling in proportion to the negative classification loss, rather than minimizing it like in REINFORCE. This may have benefits in the exploratory behavior of the sampler, as it encourages the training of diverse sets of nodes, instead of only the single best set of nodes.

\paragraph{Memory complexity.} Layer-wise sampling methods like GRAPES, FastGCN, and LADIES have a sampling space complexity of $O(Dbk)$, where $D$ is the maximum node degree, $b$ is the batch size, $k$ is the sample size, and $L$ is the number of layers. In GAS time complexity is $O(bk)$ due to the use of historical embeddings, and memory complexity is $O(DbL + N)$ where $N$ is the extra overhead for storing historical embeddings for the $N$ nodes in the graph, which can be significantly larger than $DbL$.

\begin{algorithm}[t]
\caption{One \myMethod{} epoch}\label{algo1}
\begin{algorithmic}[1]
\Require Graph $\mathcal{G}$, node embeddings $X$, node labels $Y$, target nodes $\mathcal{V}^t$, batch size $b$, sample size $k$, GCN for classification $\operatorname{GCN}_{\operatorname{C}}$, and GCN for sampling policy $\operatorname{GCN}_{\operatorname{S}}$.
\State {Divide target nodes $\mathcal{V}^t$ into batches $\mathcal{V}^{(0)}$ of size $b$}
\For {each batch $\mathcal{V}^{(0)}$}
    \State{$K^{(0)}\leftarrow \mathcal{V}^{(0)}$}
    % \State{Build adjacency matrix $\tilde{A}^{(0)}$ from $K^{(0)}$}
    \For {layer $l=1$ to $L$}
        \For {node $i$ in $\mathcal{N}(K^{(l-1)})$}
        % \State {$\mathcal{V}_n^{(l)} \gets \mathcal{N}(K^{(l-1)})$ \Comment{Get all $n$ neighbors of $K^{(l-1)}$} }
        \State {$p_i \gets \operatorname{GCN}_{S}(K^{(l-1)})_i$ \Comment{Compute probabilities of inclusion} }
        \State {$\epsilon_i\sim \operatorname{Gumbel}(0, 1)$ \Comment{Sample Gumbel noise}}
        \EndFor
        \State {$\mathcal{V}^{(l)}\gets \topk_{i \in \mathcal{N}(K^{(l-1)})} \log p_i+\epsilon_i$ \Comment{Get $k$ best nodes (Eq. \ref{gumbel})}}
        \State{$K^{(l)}\leftarrow\mathcal{V}^{(0)} \cup \mathcal{V}^{(l)}$ \Comment{Add target nodes}}
        % \State{Build adjacency matrix $\tilde{A}^{(l)}$ from $K^{(l)}$}
    \EndFor
    % \State {Pass $\mathcal{V}^ {(0)}, \ldots, \mathcal{V}^ {(L)}$ to $\operatorname{GCN}_{\operatorname{C}}$ and obtain classification loss $\mathcal{L}_{\operatorname{C}}$ given $\{(X_j,Y_j)\}_{j=1}^B$}
    \State {$\ell_{\operatorname{C}} \leftarrow \mathcal{L}_{\operatorname{C}}(X,Y, K^ {(0)},  \ldots, K^ {(L)})$ \Comment{Compute classification loss}}
    % \State {$R(A^{(0)},\ldots,A^ {(L)}) \leftarrow \exp (-\alpha \cdot \ell)$ \Comment{Calculate reward}}
    \State Compute sampling policy loss $\ell_{\operatorname{S}}$ from Eq. \ref{eq:rl} or \ref{eq:gfn}
    \State {Update parameters of $\operatorname{GCN}_{\operatorname{S}}$ by minimizing $\ell_{\operatorname{S}}$}
    \State {Update parameters of $\operatorname{GCN}_{\operatorname{C}}$ by minimizing $\ell_{\operatorname{C}}$}
\EndFor

\end{algorithmic}
\end{algorithm}

\section{Theoretical Analysis} \label{theoretical analysis}
%Our experimental results indicate that in some heterophilous graphs, \myMethod{} outperforms the baselines. 
In this section, we provide theoretical insights on the performance difference between adaptive and non-adaptive samplers, and we prove that adaptive sampling can achieve higher accuracy than non-adaptive sampling for a specific category of heterophilous graphs. We show that if there exists specific crucial information among the neighbors, there are adaptive sampling methods that perfectly identify these neighbors, while non-adaptive sampling methods cannot distinguish them from the other neighbors.  

Let $\mathcal{G}=(\mathcal{V,E)}$ be an undirected graph with a set of N nodes $\mathcal{V}=\{1, \dots, N\}$ and a set of edges $\mathcal{E}=\{e_{ij}\}_{i,j=1}^N$ where $e_{ij} \text{ denotes an edge between } v_i \text{ and } v_j$. Let $X \in \mathbb{R}^{N \times F}$ be the node features and let $Y=\{y_i\}_{i=1}^N$ be a set of node labels. The adjacency matrix $A\in \{0, 1\}^{N\times N}$ indicates a connection between a pair of nodes. Let $\mathcal{N}(v_i)=\{v_j|e_{ij} \in \mathcal{E}\}$ indicate the set of $v_i$'s neighbors and $\mathcal{N}^L(v_i)=\{v_j|l(v_i,v_j)\leq L\}$ indicate the $L$-hop neighborhood of $v_i$, where $l(v_i, v_j)$ indicate the shortest path between $v_i$ and $v_j$. In the following definitions and theorems, for simplicity, we consider the batch size of one. 

\begin{definition}[Neigbor sampling]
    A \emph{neighbor sampling method} is a method that assigns scores to neighbors of a target node, and for a given sampling budget $K$, samples $K$ neighbors proportional to their score. An $L$-layer neighbor sampling method, for each layer $l$, assigns scores to the l-hop neighbors of a target node.
\end{definition}

\begin{definition}[Featureless and feature-based sampling]
An \emph{$L$-layer feature-less sampling} method is a neighbor sampling method in which the sampling score for each node $v_i$ is a function of only its $L$-hop neighborhood, i.e. $s(v_i)=g(v_i, \mathcal{N}^L(v_i))$, where $s(v_i)$ is the sampling for node $v_i$ and $g$ is the sampling function. In an \emph{$L$-layer feature-based sampling} method, the sampling function is a function of the node features in addition to the $L$-hop neighborhood, that is, $s(v_i)=g'(v_i, X, \mathcal{N}^L(v_i))$. 
\end{definition}

In practice, all feature-based sampling methods are adaptive since the sampling methods learns the sampling score from the features, and all non-adaptive methods are featureless. Therefore, in this section, we refer to the sampling methods as adaptive and non-adaptive.  

Random sampling is a non-adaptive sampler with a constant sampling score. LADIES and FastGCN are non-adaptive samplers whose sampling scores are proportional to the node degrees. GraphSAINT-RW, i.e., the random walk-based sampler of GraphSAINT, which we used in our experiments, is a non-adaptive sampler where the score correlates with how frequently a node appears in $L$-hop neighborhoods, influenced by both the node's degree and its position in the graph. AS-GCN, PASS, and GRAPES are adaptive samplers.  

\begin{definition}[\cite{zhu2020beyond}]
    Edge homophily is the ratio of edges that connect two nodes of the same label:
    \[h=\frac{|\{e_{ij} \in \mathcal{E}: y_i=y_j\}|}{|\mathcal{E}|}\]
\end{definition}

In the next theorem, we show that the optimal adaptive and non-adaptive samplers perform differently for a certain category of graphs, with adaptive samplers perfectly sampling the influential neighbors and non-adaptive samplers sampling such nodes with a vanishing probability. 

\begin{theorem}\label{theo1}
There exist undirected graphs such that a GCN with an $L$-layer adaptive sampler and a sampling rate of $K$ neighbors per node per layer can perform with perfect accuracy, while GCNs with a non-adaptive sampler can achieve an accuracy higher than chance level only with probability $p\leq\frac{LK}{N}$. 
\end{theorem}

For the proof of this theorem, please refer to Appendix \ref{proof}. In the proof, we construct a family of distributions over graphs where the presence of a single neighbor is essential for a correct classification of a target node. 
Then, we show that only feature-based samplers, by learning from the node features, can sample those neighbors, while featureless samplers, with overwhelming probability, cannot. Since, in practice, all feature-based samplers are adaptive and, inversely, all non-adaptive samplers are featureless, we conclude that adaptive models can solve this task, while non-adaptive methods cannot. 

Additionally, we show that such graphs are heterophilous. Moreover, often, multi-label graphs are heterophilous because neighboring nodes are less likely to share the exact same set of labels. Even slight variations in the labels of neighboring nodes can lead to heterophily. Therefore, adaptive sampling methods outperform non-adaptive ones in such graphs.

\section{Experiments}\label{sec2}
Our experiments aim at answering the following research question: \emph{given a fixed sampling budget and GNN architecture, what is the effect of training with an adaptive policy for layer-wise sampling, in comparison with the related work?} While several works in the literature of sampling for GNNs have focused on classification benchmarks to demonstrate the performance of sampling algorithms, several confounding factors in their experimental setup prevent a proper understanding of whether the perceived performance improvements result from the sampling method itself. Examples of such confounding factors found in related work include using different architectures, like the GCN~\citep{kipf2016semi} or GAT~\citep{velickovic2017graph}, different sizes for the hidden layers, number of layers, batch sizes, number of nodes sampled per layer, and the number of training epochs. These are factors independent of sampling algorithms that nonetheless affect the performance in the benchmarks. We present a detailed overview of differences in the experimental setup of the related work in Table~\ref{tab:diff-experiment-setup} in the appendix.

To better understand the effect of sampling algorithms, we thus carry out experiments on a fixed GCN architecture~\citep{kipf2016semi} with two layers, a hidden size of 256, batch size of 256, sample size of 256 nodes per layer, and a fixed number of epochs per dataset (detailed in Appendix~\ref{appendix-experimental}). Under this setting, we compare \myMethod{} with the following baselines: a Random baseline that uses the same setup as \myMethod{} but with uniform inclusion probabilities; FastGCN \citep{fastGCN}, LADIES \citep{LADIES}, GraphSAINT \citep{graphsaint}, GAS \citep{fey2021gnnautoscale}, AS-GCN \citep{AS-GCN}, and PASS \cite{PASS}. With this setup, we aim to control our experiments in a way such that variations in performance can only be attributed to the sampling method.

For all baselines (except Random), we rely on their publicly available implementations. For all methods, we optimize the learning rate using the performance on the validation set. \myMethod{} requires selecting two additional hyperparameters: the learning rate for the sampling policy and, for \myMethod{}-GFN, the reward scaling parameter $\alpha$. We tune these based on the classification performance on the validation set. These hyperparameters are specific to our sampling method and do not explicitly affect the learning capacity of the GCN classifier. We refer the reader to Appendix \ref{appendix-experimental} for more details on hyperparameter settings.

\subsection{Datasets}
Most methods in the literature of sampling in GNNs are evaluated on \emph{homophilous} graphs (where the connected nodes are likely to have similar labels) and multi-class classification, where nodes are classified into one of several classes (in contrast to multi-label). To further understand the effect of sampling, we carry out experiments with \emph{heterophilous} graphs, where nodes that are connected differ in their features and labels, and multi-label datasets in which a node can be assigned one or more labels. As shown in the literature \citep{zhu2020beyond, zhao2023multi}, we argue that node classification with GCN is more challenging on these datasets compared to multi-class classification on homophilous graphs, where an adaptive policy could be beneficial. In particular, we run experiments with the following datasets for the node classification task: 
\begin{itemize}
    \item \textbf{Homophilous} graphs: citation networks (Cora, Citeseer, Pubmed with the ``full'' split) \citep{sen2008collective, yang2016revisiting}, Reddit \citep{hamilton2017inductive}, ogbn-arxiv and ogbn-products \citep{hu2020open}, and DBLP \citep{zhao2023multi}.
    \item  \textbf{Heterophilous} graphs: Flickr \citep{graphsaint}, Yelp \citep{graphsaint}, ogbn-proteins \citep{hu2020open}, BlogCat \citep{zhao2023multi}, and snap-patents \citep{snapnets}.
\end{itemize}
We included the \emph{edge homophily} ratio~\citep{zhu2020beyond} of all the datasets in Table \ref{tab1} and \ref{tab2}. For the multi-label datasets, we followed the same framework in \citep{zhao2023multi} to calculate their label homophily ratio.  Statistics about the datasets can be found in Appendix~\ref{app:data_statistics}.

\begin{table}[]
\caption{F1-scores (\%) for different sampling methods trained on \textbf{homophilous} graphs, for a batch size of 256, and sample size of 256 per layer. We report the mean and standard deviation over 10 runs. The best values among the sampling baselines (all except GAS) are in \textbf{bold}, and the second best are \underline{underlined}. MC stands for multi-class and ML stands for multi-label classification. OOM indicates out-of-memory.
% All methods were allocated the same computational resources.
% OOM stands for ``out-of-memory", which signifies that the sampling process required more GPU memory than initially allocated.
}
\label{tab1}
\centering
\resizebox{\textwidth}{!}{%\Huge
\begin{tabular}{lccccccc}
\toprule

\bf Dataset &              \bf Cora         &           \bf Citeseer        &           \bf Pubmed          &        \bf Reddit     & \bf ogbn-arxiv & \bf ogbn-products & \bf DBLP                        \\
 Homophily & $h=0.81$ & $h=0.74$ & $h=0.80$ & $h=0.78$ & $h=0.65$ & $h=0.81$ & $h=0.76$  \\
 Task & MC & MC & MC & MC & MC & MC & ML \\
\midrule                                                                                            GAS & 87.00 $\pm$ 0.19   & 85.87 $\pm$ 0.19 & 87.45 $\pm$ 0.23 & 94.75 $\pm$ 0.04 & 68.36 $\pm$ 0.55 & 74.69 $\pm$ 0.14 & 83.08 $\pm$ 0.31 \\ 
\midrule 
FastGCN                             &      76.17 $\pm$ 3.98       &      62.81 $\pm$ 7.19       &    53.52 $\pm$ 28.48 &   62.93 $\pm$ 3.28 & 39.49 $\pm$ 8.04   & 66.09 $\pm$ 3.04 & 62.93 $\pm$ 3.28    \\
LADIES                      & 76.02 $\pm$ 11.69   &  63.48 $\pm$ 12.21  & 72.81 $\pm$ 17.67      &  59.30 $\pm$ 2.69   &  43.52 $\pm$ 8.03  & 68.08 $\pm$ 1.95  & 59.97 $\pm$ 10.45               \\
GraphSAINT             &            87.28  $\pm$  0.49 &            77.28  $\pm$  0.67 &            87.45  $\pm$  0.75 &            91.47 $\pm$ 0.94       &          \underline{63.54} $\pm$ 1.75 &      67.66 $\pm$ 0.69  &  77.09 $\pm$ 0.56                         \\
AS-GCN                 & 85.60 $\pm$ 0.54 &\bf 79.21 $\pm$ 0.19 & \bf 90.58 $\pm$ 0.40 & 93.52 $\pm$ 0.40 & \bf 65.38 $\pm$ 1.80 & \bf 73.94 $\pm$ 0.40 & \bf83.24 $\pm$ 0.51       \\
PASS                & 82.03 $\pm$ 0.074 & 77.17 $\pm$ 0.69 & 88.33 $\pm$ 0.45 & OOM &  58.32 $\pm$ 0.05 & OOM & 54.40 $\pm$ 2.31       \\
Random & 86.58 $\pm$ 0.33 & 78.29 $\pm$ 0.52 & \underline{90.09} $\pm$ 0.17 & \underline{94.16} $\pm$ 0.06 &61.35 $\pm$ 0.32 & 70.47 $\pm$ 0.32& 76.87 $\pm$ 0.24 \\
\midrule
\myMethod{}-RL (ours) & \bf 87.62 $\pm$ 0.48 & \underline{78.75} $\pm$ 0.35 & 89.40 $\pm$ 0.34 & 94.09 $\pm$ 0.05 & 62.58 $\pm$ 0.64 &\underline{71.45} $\pm$ 0.20 & 76.88 $\pm$ 0.33 \\
\myMethod{}-GFN (ours) & \underline{87.29} $\pm$ 0.32 & \underline{78.75} $\pm$ 0.33 & 89.46 $\pm$ 0.34     & \bf 94.30 $\pm$ 0.06     & 61.86 $\pm$ 0.51 &  70.66 $\pm$ 0.30  & \underline{77.14} $\pm$ 0.48                  \\
\midrule
Rank & 1 & 2 & 3 & 1 & 3 & 2 & 2 \\
\bottomrule
\end{tabular}
}
\end{table}

\begin{table}[]%\Huge
\caption{F1-scores (\%) for different sampling methods trained on \textbf{heterophilous} graphs for a batch size of 256, and a sample size of 256 per layer. We report the mean and standard deviation over 10 runs. The best values among the sampling baselines (all except GAS) are in \textbf{bold}, and the second best are \underline{underlined}. MC stands for multi-class and ML stands for multi-label classification. OOM indicates out of memory.}
\label{tab2}
\centering
\resizebox{\textwidth}{!}{
\begin{tabular}{lccccc}
\toprule
\bf Dataset             &              \bf Flickr      &      \bf snap-patents     &           \bf Yelp        &           \bf ogbn-proteins          &        \bf BlogCat                       \\
 Homophily & $h=0.31$ & $h=0.22$ & $h=0.22$ & $h=0.15$ & $h=0.1$ \\
 Task & MC & MC & ML & ML & ML  \\
\midrule                                                                                                                                                              GAS & 49.96 $\pm$ 0.28 & 38.04 $\pm$ 0.20 & 37.81 $\pm$ 0.07 & 7.55 $\pm$ 0.01 & 6.07 $\pm$ 0.04  \\
\midrule                
FastGCN                & 46.40 $\pm$ 2.51 & 29.68 $\pm$ 0.61 &  29.29 $\pm$ 4.39  &   7.80 $\pm$ 0.55   & 6.44 $\pm$ 0.41  \\
LADIES                 & 47.19 $\pm$ 3.42    & 29.09 $\pm$ 1.88 & 18.92 $\pm$ 3.18 & 4.31 $\pm$ 0.11 & 6.74 $\pm$ 0.67   \\
GraphSAINT             &      48.01 $\pm$ 1.44 & 28.01 $\pm$ 0.57 &        34.68 $\pm$ 0.70   & 9.94 $\pm$ 0.07 &  6.89 $\pm$ 0.96    \\
AS-GCN                 & 48.42 $\pm$ 1.20 & \bf 31.04 $\pm$ 0.19 & 38.51 $\pm$ 1.45 & 5.20 $\pm$ 0.36 & 5.43 $\pm$ 0.54   \\
PASS                & 44.49 $\pm$ 0.76 & OOM & OOM & 7.71 $\pm$ 0.12 & 6.88 $\pm$ 0.59   \\
Random                 & \underline{49.39} $\pm$ 0.23   & \underline{29.74} $\pm$ 0.34 & 40.63 $\pm$ 0.13 & 10.82 $\pm$ 0.05 &     7.13 $\pm$ 0.97    \\
\midrule
\myMethod{}-RL (ours)             & \bf49.54 $\pm$ 0.67 & 29.16 $\pm$ 0.54  & \underline{40.69} $\pm$ 0.55& \bf 11.78 $\pm$ 0.14 &\underline{9.06} $\pm$ 0.68   \\
\myMethod{}-GFN (ours) &    49.29  $\pm$  0.32   & 29.58 $\pm$ 0.24     &   \bf  44.57 $\pm$ 0.88  &  \underline{11.57} $\pm$ 0.18   &   \bf 9.22 $\pm$ 0.40   \\
\midrule 
Rank & 1 & 4 & 1 & 1 & 1 \\
\bottomrule
\end{tabular}
}
\end{table}

\subsection{Results}
\label{section:results}

\textbf{Comparison with sampling methods} We present the F1 scores of GCNs trained via sampling for \myMethod{} and the sampling-based baselines in Table~\ref{tab1} and~\ref{tab2}.
We observe that on the majority of heterophilous datasets, and all the heterophilous multi-label datasets, both variations of \myMethod{} achieve the highest F1-score.
%Unlike the other methods, \myMethod{} ranks consistently within the top 3 methods, and its average rank is the highest, 
As we mentioned earlier, node classification on heterophilous graphs is a challenging task for GCNs, due to the diversity of the neighbor nodes' labels. Our results suggest that \myMethod{}' ability to adapt the sampling policy to particular features of the data allows \myMethod{} to learn the complex patterns across the heterophilous datasets. Table \ref{tab1} shows that on most homophilous graphs, AS-GCN achieves the highest F1-score, indicating the importance of adaptive sampling. However, unlike \myMethod{}, this method is only adaptive to the sampling variance and fails to outperform \myMethod{} on the heterophilous graphs. Moreover, our results show that PASS, another adaptive method, is unable to compete with \myMethod{} and results in an out-of-memory error on several large datasets. This is due to its node-wise sampling feature, which results in a disconnected sampled graph.

FastGCN and LADIES, which use a fixed policy, fail to compete with the other methods in our experimental setup. They assign a higher probability to nodes with higher degrees. This heuristic results in neglecting informative low-degree nodes. Although outperforming LADIES and FastGCN, GraphSAINT, which also uses a fixed sampling policy, fails to outperform the more competitive baselines. We observe that on some datasets, Random sampling achieves a comparable F1-score to \myMethod{} and the baselines. This mainly happens on simple datasets, such as homophilous graphs, indicating that on these graphs, a simple sampling approach is sufficient.

\begin{wrapfigure}[17]{r}{0.4\textwidth}
    \includegraphics[width=0.4\textwidth]{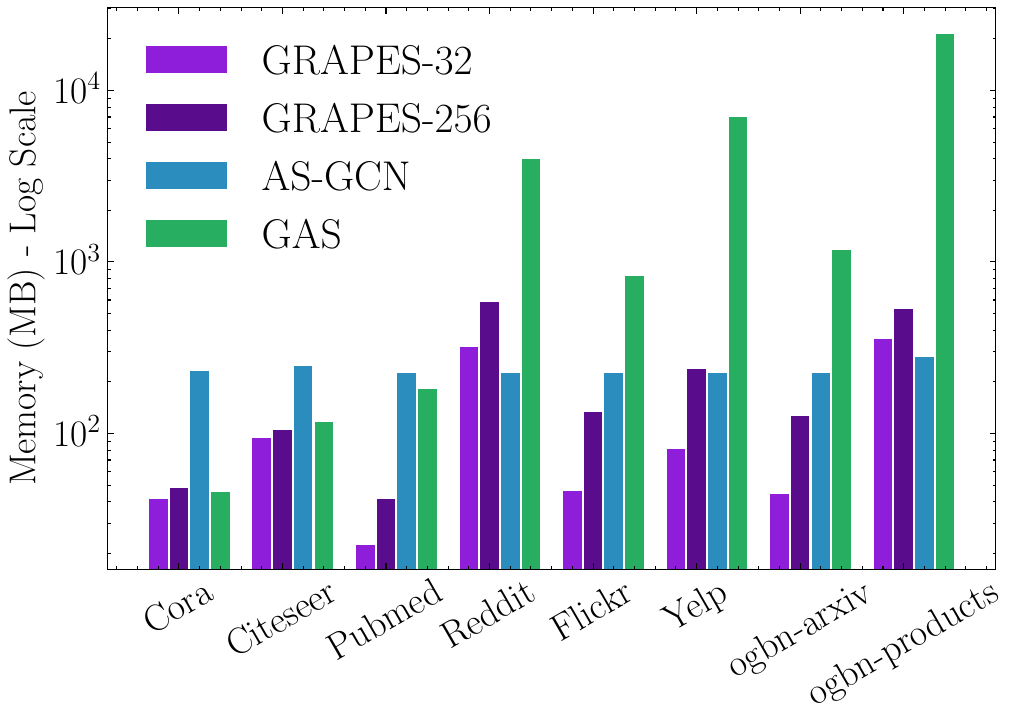}
    \caption{GPU peak memory allocation (MB) for GAS, and GRAPES-GFN-32 and GRAPES-GFN-256.}
    \label{fig:memory-gas-grapes}
\end{wrapfigure}

\textbf{Comparison with a non-sampling scalable method} GAS \citep{fey2021gnnautoscale} is a non-sampling method that uses the historical embeddings of the 1-hop neighbors of the target nodes. We present classification results in Table~\ref{tab2}, and we visualize memory usage in Fig.~\ref{fig:memory-gas-grapes}. While GAS has higher classification F1-scores for some datasets, \myMethod{} achieves comparable F1-scores and significantly outperforms GAS for Yelp, ogbn-proteins, and BlogCat, all heterophilous multi-label graphs. Once again, these results indicate the effectiveness of the adaptivity of \myMethod{} in sampling influential nodes in complex graphs. For the memory usage, we compare GAS with AS-GCN and the GFlowNet version of \myMethod{}-32 and \myMethod{}-256, where $32$ and $256$ (the default value) indicate the sample size per layer.
The memory usage results show that, even with a large sample size, \myMethod{} can use up to an order of magnitude less GPU memory than GAS, especially for large datasets. In large, densely connected graphs such as Reddit, the 1-hop neighborhood can be massive.
Then, the difference in memory use for \myMethod{}, which only sees a small set of neighbors, and GAS, which uses all the neighbors, is significant.
While GAS occasionally achieves higher F1-scores, it consistently demands more memory, indicating a potential compromise between accuracy and computational efficiency. In contrast, \myMethod{} strikes a balance, delivering comparable F1-score with more modest memory footprints. In most cases, \myMethod{} also uses less memory than AS-GCN, another sampling-based baseline. In other cases, \myMethod{} uses more memory than AS-GCN, which we attribute to the fact that there is an additional GNN in \myMethod{} for learning the sampling policy. Some variations are due to implementation details\footnote{We used PyTorch for our implementation, whereas AS-GCN is implemented in Tensorflow, which manages GPU memory differently.}, though in general we observe a significant advantage of layer-wise sampling methods over historical embeddings.

% \begin{figure}[t]
%     \centering
%     \includegraphics[width=0.5\linewidth]{figures/memory-gas-grapes-plot.pdf}
%     \caption{Caption}
%     \label{fig:enter-label}
% \end{figure}

\begin{figure}[t]
    \centering
    \includegraphics[width=\textwidth]{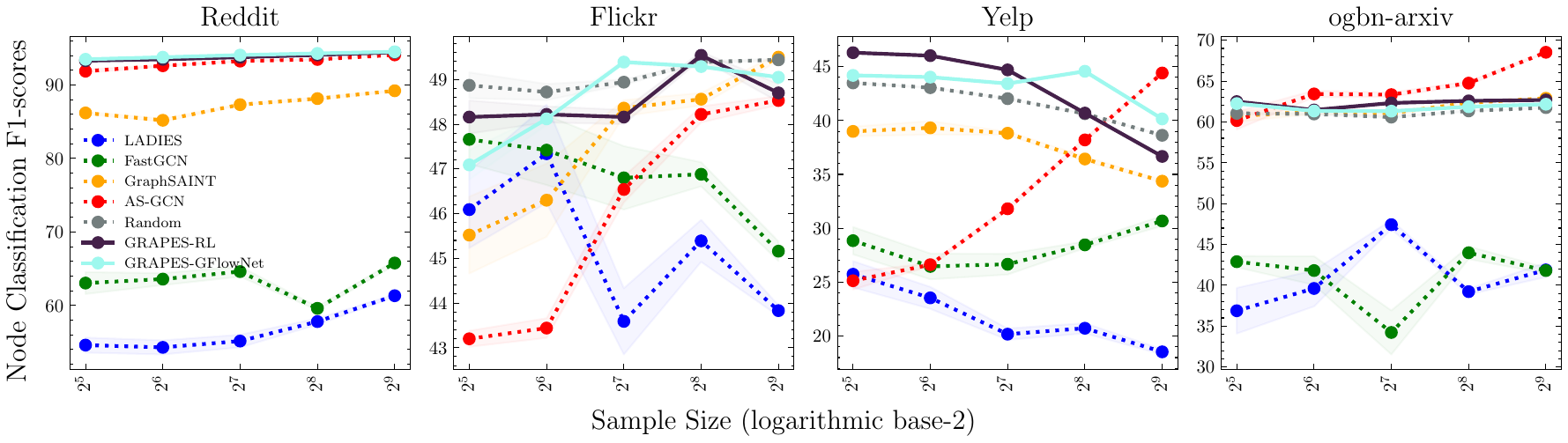}
    \caption{Comparative analysis of classification accuracy across different sampling sizes for sampling baseline and \myMethod{}. 
    % The x-axis, in a logarithmic scale (base-2), represents the number of samples sampled by the sampling methods. The y-axis denotes the classification F1-scores in percentage (\%). 
    We repeated each experiment five times: The shaded regions show the 95\% confidence intervals. 
    %GAS has been excluded from this plot because it does not sample nodes but instead always sees the entire graph. 
    }
    \label{fig:sampling-effect}
\end{figure}

\noindent \textbf{\myMethod{}\ is robust to low sample sizes.}
A desirable property of sampling methods, in contrast with full-batch GNNs or methods like GAS (which relies on historical embeddings), is the ability to control the sample size to reduce memory usage as needed. To study this property, we show in Figure \ref{fig:sampling-effect} the effects of varying the sample size on Reddit, Flickr, Yelp, and ogbn-arxiv. Our results show that both the RL and GFN variants of \myMethod{} are robust to low sample size, and achieve strong performance with fewer sampled nodes needed than the baselines, enabling training GCNs on larger graphs while using less GPU memory. Random sampling also exhibits robustness to sample size for all datasets except Yelp, where accuracy drops in larger sample sizes. \myMethod{}-RL shows the same behavior while performing slightly better than Random on Yelp.
%The drop in accuracy can happen because the node features are informative enough for training. Therefore, adding a large set of neighbors in Yelp adds more noise to the graph, negatively affecting the accuracy.
AS-GCN and GraphSAINT show the largest dependence on sample size, especially on Flickr and Yelp.

\begin{figure}[t]
    \centering
    \includegraphics[width=\textwidth]{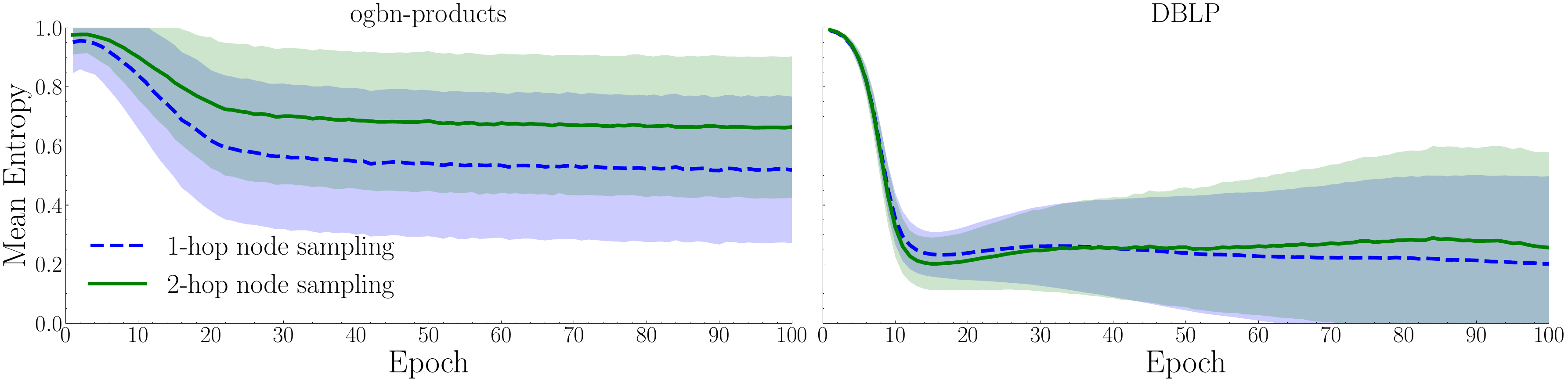} 
    \caption{Entropy for the ogbn-products and DBLP datasets. The mean is the entropy in bits of the node probability, averaged over nodes. The shaded region indicates the standard deviation of the entropy over all nodes.}
    \label{fig:mean_entropy_comparison}
\end{figure}

\noindent \textbf{\myMethod{}\ learns strong preferences over nodes.} The ability to selectively choose \emph{influential} nodes is a crucial property of \myMethod{}.
Figure \ref{fig:mean_entropy_comparison} shows the mean and standard deviation of base 2 entropy for the node preference probabilities for the two layers of $\operatorname{GCN}_{\operatorname{S}}$ for ogbn-products and DBLP for \myMethod{}-GFN. The probabilities show preference towards particular nodes with a Bernoulli distribution. 
A well-trained model must have a high preference (probability close to 1) for some nodes and a low preference (probability close to 0) for the rest. Therefore, we would like a low average entropy with a high standard deviation. As the figure shows, the mean entropy in both layers decreases from almost 1 and converges to a value above 0, while the standard deviation increases. This indicates that for ogbn-products, the sampling policy initially assigned a probability near $0.5$, indicating little preference. However, after several training epochs, \myMethod{} starts preferring some nodes, resulting in lower mean entropy. We observe similar behavior for most datasets (DBLP, BlogCat, Yelp, ogbn-arxiv, and ogbn-proteins). However, we observe that the sampling policy exhibits no preferences among the nodes for the other datasets. For more details about the other datasets, see Appendix \ref{appendix-entropy}. 
% Moreover, as shown in Appendix \ref{appendix-entropy} we observe that in the last epoch for Yelp, \myMethod{} tends to prefer nodes with fewer labels. This means that in a multi-label classification task, sampling nodes with fewer labels results in a lower loss, possibly because learning fewer labels can be easier for the GNN. We leave further investigation in this direction to future work.

\section{Known Limitations and Future Work}
Our experimental results show that \myMethod{} outperforms the baselines on some, but not all, heterophilous datasets. We provide a theoretical analysis that shows the effectiveness of adaptive sampling on certain types of heterophilous graphs. Further analysis is required to understand which graph characteristics contribute to the performance gain observed with adaptive sampling methods such as \myMethod{}.  

Moreover, in our experiments, we focused only on the problem of node classification. However, \myMethod{} is not tied to a particular downstream task. \myMethod{} assumes access to a tractable reward function \citep{bengio2021gflow-fundation}.
Therefore, \myMethod{} will be applicable for other graph-related tasks, like link prediction and unsupervised representation learning. 

Additionally, there is room for a more thorough investigation of the properties of the subgraphs that \myMethod{} samples. The experiments at the end of Section~\ref{section:results} and in Appendix \ref{appendix-entropy} are a first step towards this. We leave these directions for future work. 

Another direction for future work is applying \myMethod{} to architectures designed to address heterophily, such as the ones proposed by \citet{zhu2020beyond, abu2019mixhop}. By leveraging \myMethod{}, these architectures could potentially overcome their scalability issues.

\section{Discussion, Broader Impacts, and Conclusion}
\label{discussion}
We propose \myMethod{}, an adaptive graph sampling method based on reinforcement learning and GFlowNet, facilitating the scalability of training GNNs on massive graphs. \myMethod{} samples a subgraph of influential nodes by learning node preferences that adapt to classification loss, which depends on node features, GNN architecture, classification task, and graph topology.
% \myMethod{} learns the importance of the nodes by adapting to the classification loss. We show how a sampling policy can build a subgraph of influential nodes by learning node preferences that adapt to node features, GNN architecture, classification task, and graph topology. 
Our experiments demonstrate that \myMethod{} effectively selects nodes from large-scale graphs and
% achieves consistent performance over state-of-the-art sampling methods, especially on complex graph structures such as heterophilous graphs.
achieves state-of-the-art performance in multi-label classification tasks on heterophilous graphs, while performing competitively for multi-class classification on homophilous graphs.
Compared to the other sampling methods, \myMethod{} can maintain high classification accuracy even with lower sample sizes, indicating \myMethod{}' ability to scale to larger graphs by sampling a small but influential set of nodes. \myMethod{} achieves comparable performance to GAS, while using up to an order of magnitude less memory. 

\textbf{Focus on Heterophily in Sampling.} Previous works have overlooked the impact of sampling on heterophilous and multi-label graphs. To our knowledge, this is the first work to compare sampling effects on both homophilous and heterophilous graphs, where adaptive sampling is especially crucial due to the diversity among neighbors, making node selection vital for GNN accuracy.

\textbf{Lack of Uniform Evaluation Protocol.} Existing methods in the literature report performance on graph sampling under settings with different GCN architectures, regularization techniques, feature normalization strategies, and data splits, among other differences. These differences made it challenging to determine the benefits of each sampling method. This motivated us to implement a unified protocol across all methods, where we keep the architecture fixed. 
We encourage future work to consider a similar methodology for a fair evaluation, or an experimental review study, as is common in other areas of machine learning research on graphs~\citep{shchur2018pitfalls,ruffinelli2019you}.

\textbf{Broader Impacts.} Our paper focuses on fundamental research to understand and advance sampling in large-scale graphs. There may be multiple potential societal consequences of our work, but there are none that can be easily predicted and specifically highlighted.

\section*{Acknowledgements}
Taraneh Younesian was funded by Huawei DREAMS Lab. All content represents the opinion of the authors, which is not necessarily shared nor endorsed by their respective employers and/or sponsors in Huawei DREAMS Lab. Daniel Daza was partially funded by Elsevier's Discovery Lab. 
Emile van Krieken was funded by ELIAI (The Edinburgh Laboratory for Integrated Artificial Intelligence), EPSRC (grant no. EP/W002876/1).
We thank Michael Cochez and Ruud van Bakel for insightful discussions. 

% \noindent \textbf{Known Limitations and Future Work.}
% In our experiments, we only focused on the problem of node classification. However, \myMethod{} is not tied to a particular downstream task. \myMethod{} assumes access to a tractable reward function \citep{bengio2021gflow-fundation}.
% Therefore, GRAPES will be applicable for other graph-related tasks, like link prediction and unsupervised representation learning. Another limitation is that we only evaluate \myMethod{} on a GCN architecture with a fixed number of layers. Additionally, there is room for a more thorough investigation into the properties of the subgraphs that GRAPES samples. The experiments at the end of Section~\ref{section:results} are a first step towards this. We leave these directions for future work. Another direction for future work is applying GRAPES to architectures designed to address heterophily, such as the ones proposed by \citet{zhu2020beyond, abu2019mixhop}. By leveraging GRAPES, these architectures could potentially overcome their scalability issues.

\bibliography{main}

\begin{thebibliography}{59}
\providecommand{\natexlab}[1]{#1}
\providecommand{\url}[1]{\texttt{#1}}
\expandafter\ifx\csname urlstyle\endcsname\relax
  \providecommand{\doi}[1]{doi: #1}\else
  \providecommand{\doi}{doi: \begingroup \urlstyle{rm}\Url}\fi

\bibitem[Abu-El-Haija et~al.(2019)Abu-El-Haija, Perozzi, Kapoor, Alipourfard, Lerman, Harutyunyan, Ver~Steeg, and Galstyan]{abu2019mixhop}
Sami Abu-El-Haija, Bryan Perozzi, Amol Kapoor, Nazanin Alipourfard, Kristina Lerman, Hrayr Harutyunyan, Greg Ver~Steeg, and Aram Galstyan.
\newblock Mixhop: Higher-order graph convolutional architectures via sparsified neighborhood mixing.
\newblock In \emph{international conference on machine learning}, pp.\  21--29. PMLR, 2019.

\bibitem[Abu-El-Haija et~al.(2023)Abu-El-Haija, Dillon, Fatemi, Axiotis, Bulut, Gasteiger, Perozzi, and Bateni]{abu2023submix}
Sami Abu-El-Haija, Joshua~V Dillon, Bahare Fatemi, Kyriakos Axiotis, Neslihan Bulut, Johannes Gasteiger, Bryan Perozzi, and Mohammadhossein Bateni.
\newblock Submix: Learning to mix graph sampling heuristics.
\newblock In \emph{Uncertainty in Artificial Intelligence}, pp.\  1--10. PMLR, 2023.

\bibitem[Ahmed et~al.(2023)Ahmed, Zeng, Niepert, and den Broeck]{ahmed2023simple}
Kareem Ahmed, Zhe Zeng, Mathias Niepert, and Guy~Van den Broeck.
\newblock {SIMPLE}: A gradient estimator for k-subset sampling.
\newblock In \emph{The Eleventh International Conference on Learning Representations}, 2023.

\bibitem[Anis et~al.(2016)Anis, Gadde, and Ortega]{anis2016efficient}
Aamir Anis, Akshay Gadde, and Antonio Ortega.
\newblock Efficient sampling set selection for bandlimited graph signals using graph spectral proxies.
\newblock \emph{IEEE Transactions on Signal Processing}, 64\penalty0 (14):\penalty0 3775--3789, 2016.

\bibitem[Bengio et~al.(2021{\natexlab{a}})Bengio, Jain, Korablyov, Precup, and Bengio]{bengio2021flow}
Emmanuel Bengio, Moksh Jain, Maksym Korablyov, Doina Precup, and Yoshua Bengio.
\newblock Flow network based generative models for non-iterative diverse candidate generation.
\newblock \emph{Advances in Neural Information Processing Systems}, 34:\penalty0 27381--27394, 2021{\natexlab{a}}.

\bibitem[Bengio et~al.(2013)Bengio, L{\'{e}}onard, and Courville]{bengio2013st}
Yoshua Bengio, Nicholas L{\'{e}}onard, and Aaron~C. Courville.
\newblock Estimating or propagating gradients through stochastic neurons for conditional computation.
\newblock \emph{CoRR}, abs/1308.3432, 2013.
\newblock URL \url{http://arxiv.org/abs/1308.3432}.

\bibitem[Bengio et~al.(2021{\natexlab{b}})Bengio, Lahlou, Deleu, Hu, Tiwari, and Bengio]{bengio2021gflow-fundation}
Yoshua Bengio, Salem Lahlou, Tristan Deleu, Edward~J Hu, Mo~Tiwari, and Emmanuel Bengio.
\newblock Gflownet foundations.
\newblock \emph{arXiv preprint arXiv:2111.09266}, 2021{\natexlab{b}}.

\bibitem[Biewald(2020)]{wandb}
Lukas Biewald.
\newblock Experiment tracking with weights and biases, 2020.
\newblock URL \url{https://www.wandb.com/}.
\newblock Software available from wandb.com.

\bibitem[Chen et~al.(2018{\natexlab{a}})Chen, Zhu, and Song]{vr-gcn}
Jianfei Chen, Jun Zhu, and Le~Song.
\newblock Stochastic training of graph convolutional networks with variance reduction.
\newblock In \emph{International Conference on Machine Learning}, pp.\  942--950. PMLR, 2018{\natexlab{a}}.

\bibitem[Chen et~al.(2018{\natexlab{b}})Chen, Ma, and Xiao]{fastGCN}
Jie Chen, Tengfei Ma, and Cao Xiao.
\newblock {FastGCN}: Fast learning with graph convolutional networks via importance sampling.
\newblock In \emph{6th International Conference on Learning Representations, {ICLR} 2018, Conference Track Proceedings}. OpenReview.net, 2018{\natexlab{b}}.

\bibitem[Chiang et~al.(2019)Chiang, Liu, Si, Li, Bengio, and Hsieh]{ClusterGCN}
Wei-Lin Chiang, Xuanqing Liu, Si~Si, Yang Li, Samy Bengio, and Cho-Jui Hsieh.
\newblock Cluster-gcn: An efficient algorithm for training deep and large graph convolutional networks.
\newblock In \emph{Proceedings of the 25th ACM SIGKDD international conference on knowledge discovery \& data mining}, pp.\  257--266, 2019.

\bibitem[Cong et~al.(2020)Cong, Forsati, Kandemir, and Mahdavi]{cong2020minimal}
Weilin Cong, Rana Forsati, Mahmut Kandemir, and Mehrdad Mahdavi.
\newblock Minimal variance sampling with provable guarantees for fast training of graph neural networks.
\newblock In \emph{Proceedings of the 26th ACM SIGKDD International Conference on Knowledge Discovery \& Data Mining}, pp.\  1393--1403, 2020.

\bibitem[Cong et~al.(2023)Cong, Shi, Li, Yang, He, and Pei]{cong2023fairsample}
Zicun Cong, Baoxu Shi, Shan Li, Jaewon Yang, Qi~He, and Jian Pei.
\newblock Fairsample: Training fair and accurate graph convolutional neural networks efficiently.
\newblock \emph{IEEE Transactions on Knowledge and Data Engineering}, 36\penalty0 (4):\penalty0 1537--1551, 2023.

\bibitem[Deleu et~al.(2022)Deleu, G{\'o}is, Emezue, Rankawat, Lacoste-Julien, Bauer, and Bengio]{deleu2022bayesian}
Tristan Deleu, Ant{\'o}nio G{\'o}is, Chris Emezue, Mansi Rankawat, Simon Lacoste-Julien, Stefan Bauer, and Yoshua Bengio.
\newblock Bayesian structure learning with generative flow networks.
\newblock In \emph{Uncertainty in Artificial Intelligence}, pp.\  518--528. PMLR, 2022.

\bibitem[Fey \& Lenssen(2019)Fey and Lenssen]{Fey/Lenssen/2019}
Matthias Fey and Jan~E. Lenssen.
\newblock Fast graph representation learning with {PyTorch Geometric}.
\newblock In \emph{ICLR Workshop on Representation Learning on Graphs and Manifolds}, 2019.

\bibitem[Fey et~al.(2021)Fey, Lenssen, Weichert, and Leskovec]{fey2021gnnautoscale}
Matthias Fey, Jan~E Lenssen, Frank Weichert, and Jure Leskovec.
\newblock Gnnautoscale: Scalable and expressive graph neural networks via historical embeddings.
\newblock In \emph{International conference on machine learning}, pp.\  3294--3304. PMLR, 2021.

\bibitem[Gao et~al.(2022)Gao, Fu, Sun, and Coley]{gao2022molecule}
Wenhao Gao, Tianfan Fu, Jimeng Sun, and Connor Coley.
\newblock Sample efficiency matters: a benchmark for practical molecular optimization.
\newblock \emph{Advances in Neural Information Processing Systems}, 35:\penalty0 21342--21357, 2022.

\bibitem[Geng et~al.(2023)Geng, Chen, He, Zeng, Han, Chai, and Yan]{geng2023pyramid}
Haoyu Geng, Chao Chen, Yixuan He, Gang Zeng, Zhaobing Han, Hua Chai, and Junchi Yan.
\newblock Pyramid graph neural network: A graph sampling and filtering approach for multi-scale disentangled representations.
\newblock In \emph{Proceedings of the 29th ACM SIGKDD Conference on Knowledge Discovery and Data Mining}, pp.\  518--530, 2023.

\bibitem[Hamilton et~al.(2017)Hamilton, Ying, and Leskovec]{hamilton2017inductive}
Will Hamilton, Zhitao Ying, and Jure Leskovec.
\newblock Inductive representation learning on large graphs.
\newblock \emph{Advances in Neural Information Processing Systems}, 30, 2017.

\bibitem[Hornik et~al.(1989)Hornik, Stinchcombe, and White]{hornik1989universal}
Kurt Hornik, Maxwell~B. Stinchcombe, and Halbert White.
\newblock Multilayer feedforward networks are universal approximators.
\newblock \emph{Neural Networks}, 2\penalty0 (5):\penalty0 359--366, 1989.
\newblock \doi{10.1016/0893-6080(89)90020-8}.
\newblock URL \url{https://doi.org/10.1016/0893-6080(89)90020-8}.

\bibitem[Hu et~al.(2020)Hu, Fey, Zitnik, Dong, Ren, Liu, Catasta, and Leskovec]{hu2020open}
Weihua Hu, Matthias Fey, Marinka Zitnik, Yuxiao Dong, Hongyu Ren, Bowen Liu, Michele Catasta, and Jure Leskovec.
\newblock Open graph benchmark: Datasets for machine learning on graphs.
\newblock \emph{Advances in neural information processing systems}, 33:\penalty0 22118--22133, 2020.

\bibitem[Huang et~al.(2018)Huang, Zhang, Rong, and Huang]{AS-GCN}
Wenbing Huang, Tong Zhang, Yu~Rong, and Junzhou Huang.
\newblock Adaptive sampling towards fast graph representation learning.
\newblock \emph{Advances in neural information processing systems}, 31, 2018.

\bibitem[Huijben et~al.(2022)Huijben, Kool, Paulus, and Van~Sloun]{huijben2022review-gumbel}
Iris~AM Huijben, Wouter Kool, Max~B Paulus, and Ruud~JG Van~Sloun.
\newblock A review of the gumbel-max trick and its extensions for discrete stochasticity in machine learning.
\newblock \emph{IEEE Transactions on Pattern Analysis and Machine Intelligence}, 45\penalty0 (2):\penalty0 1353--1371, 2022.

\bibitem[Jain et~al.(2022)Jain, Bengio, Hernandez-Garcia, Rector-Brooks, Dossou, Ekbote, Fu, Zhang, Kilgour, Zhang, et~al.]{jain2022biological}
Moksh Jain, Emmanuel Bengio, Alex Hernandez-Garcia, Jarrid Rector-Brooks, Bonaventure~FP Dossou, Chanakya~Ajit Ekbote, Jie Fu, Tianyu Zhang, Michael Kilgour, Dinghuai Zhang, et~al.
\newblock Biological sequence design with gflownets.
\newblock In \emph{International Conference on Machine Learning}, pp.\  9786--9801. PMLR, 2022.

\bibitem[Jain et~al.(2023)Jain, Deleu, Hartford, Liu, Hernandez-Garcia, and Bengio]{jain2023scientific}
Moksh Jain, Tristan Deleu, Jason Hartford, Cheng-Hao Liu, Alex Hernandez-Garcia, and Yoshua Bengio.
\newblock Gflownets for {AI}-driven scientific discovery.
\newblock \emph{Digital Discovery}, 2\penalty0 (3):\penalty0 557--577, 2023.

\bibitem[Kingma \& Ba(2014)Kingma and Ba]{kingma2014adam}
Diederik~P Kingma and Jimmy Ba.
\newblock Adam: A method for stochastic optimization.
\newblock \emph{arXiv preprint arXiv:1412.6980}, 2014.

\bibitem[Kipf \& Welling(2016)Kipf and Welling]{kipf2016semi}
Thomas~N Kipf and Max Welling.
\newblock Semi-supervised classification with graph convolutional networks.
\newblock \emph{arXiv preprint arXiv:1609.02907}, 2016.

\bibitem[Leskovec \& Krevl(2014)Leskovec and Krevl]{snapnets}
Jure Leskovec and Andrej Krevl.
\newblock {SNAP Datasets}: {Stanford} large network dataset collection.
\newblock \url{http://snap.stanford.edu/data}, 2014.

\bibitem[Li et~al.(2022)Li, Huang, and Zitnik]{li2022graph}
Michelle~M Li, Kexin Huang, and Marinka Zitnik.
\newblock Graph representation learning in biomedicine and healthcare.
\newblock \emph{Nature Biomedical Engineering}, 6\penalty0 (12):\penalty0 1353--1369, 2022.

\bibitem[Li et~al.(2023)Li, Li, Li, Hao, and Pang]{li2023dag}
Wenqian Li, Yinchuan Li, Zhigang Li, Jianye Hao, and Yan Pang.
\newblock Dag matters! gflownets enhanced explainer for graph neural networks.
\newblock \emph{arXiv preprint arXiv:2303.02448}, 2023.

\bibitem[Liu et~al.(2020)Liu, Wu, Zhang, Zhou, Yang, Song, and Qi]{gnn-bs}
Ziqi Liu, Zhengwei Wu, Zhiqiang Zhang, Jun Zhou, Shuang Yang, Le~Song, and Yuan Qi.
\newblock Bandit samplers for training graph neural networks.
\newblock \emph{Advances in Neural Information Processing Systems}, 33:\penalty0 6878--6888, 2020.

\bibitem[Liu et~al.(2023)Liu, Zhou, Jiang, Li, Chen, Choi, and Hu]{liu2023dspar}
Zirui Liu, Kaixiong Zhou, Zhimeng Jiang, Li~Li, Rui Chen, Soo-Hyun Choi, and Xia Hu.
\newblock Dspar: An embarrassingly simple strategy for efficient gnn training and inference via degree-based sparsification.
\newblock \emph{Transactions on Machine Learning Research}, 2023.

\bibitem[Ma et~al.(2025)Ma, Sheng, Li, Gao, Hao, Yang, Nie, Jiang, Zhang, and Cui]{ma2025acceleration}
Lu~Ma, Zeang Sheng, Xunkai Li, Xinyi Gao, Zhezheng Hao, Ling Yang, Xiaonan Nie, Jiawei Jiang, Wentao Zhang, and Bin Cui.
\newblock Acceleration algorithms in gnns: A survey.
\newblock \emph{IEEE Transactions on Knowledge and Data Engineering}, 2025.

\bibitem[Malkin et~al.(2022{\natexlab{a}})Malkin, Jain, Bengio, Sun, and Bengio]{malkin2022trajectorybalance}
Nikolay Malkin, Moksh Jain, Emmanuel Bengio, Chen Sun, and Yoshua Bengio.
\newblock Trajectory balance: Improved credit assignment in gflownets.
\newblock \emph{Advances in Neural Information Processing Systems}, 35:\penalty0 5955--5967, 2022{\natexlab{a}}.

\bibitem[Malkin et~al.(2022{\natexlab{b}})Malkin, Lahlou, Deleu, Ji, Hu, Everett, Zhang, and Bengio]{malkin2022gflownets-off}
Nikolay Malkin, Salem Lahlou, Tristan Deleu, Xu~Ji, Edward~J Hu, Katie~E Everett, Dinghuai Zhang, and Yoshua Bengio.
\newblock Gflownets and variational inference.
\newblock In \emph{The Eleventh International Conference on Learning Representations}, 2022{\natexlab{b}}.

\bibitem[Mohamed et~al.(2020)Mohamed, Rosca, Figurnov, and Mnih]{mohamedMonteCarloGradient2020}
Shakir Mohamed, Mihaela Rosca, Michael Figurnov, and Andriy Mnih.
\newblock Monte carlo gradient estimation in machine learning.
\newblock \emph{Journal of Machine Learning Research}, 21:\penalty0 132:1--132:62, 2020.

\bibitem[Nettleton(2013)]{nettleton2013data}
David~F Nettleton.
\newblock Data mining of social networks represented as graphs.
\newblock \emph{Computer Science Review}, 7:\penalty0 1--34, 2013.

\bibitem[Ruffinelli et~al.(2019)Ruffinelli, Broscheit, and Gemulla]{ruffinelli2019you}
Daniel Ruffinelli, Samuel Broscheit, and Rainer Gemulla.
\newblock You can teach an old dog new tricks! on training knowledge graph embeddings.
\newblock In \emph{International Conference on Learning Representations}, 2019.

\bibitem[Ruiz et~al.(2023)Ruiz, Chamon, and Ribeiro]{ruiz2023transferability}
Luana Ruiz, Luiz~FO Chamon, and Alejandro Ribeiro.
\newblock Transferability properties of graph neural networks.
\newblock \emph{IEEE Transactions on Signal Processing}, 2023.

\bibitem[Sen et~al.(2008)Sen, Namata, Bilgic, Getoor, Galligher, and Eliassi-Rad]{sen2008collective}
Prithviraj Sen, Galileo Namata, Mustafa Bilgic, Lise Getoor, Brian Galligher, and Tina Eliassi-Rad.
\newblock Collective classification in network data.
\newblock \emph{AI magazine}, 29\penalty0 (3):\penalty0 93--93, 2008.

\bibitem[Serafini \& Guan(2021)Serafini and Guan]{serafini2021scalable}
Marco Serafini and Hui Guan.
\newblock Scalable graph neural network training: The case for sampling.
\newblock \emph{ACM SIGOPS Operating Systems Review}, 55\penalty0 (1):\penalty0 68--76, 2021.

\bibitem[Shchur et~al.(2018)Shchur, Mumme, Bojchevski, and G{\"u}nnemann]{shchur2018pitfalls}
Oleksandr Shchur, Maximilian Mumme, Aleksandar Bojchevski, and Stephan G{\"u}nnemann.
\newblock Pitfalls of graph neural network evaluation.
\newblock \emph{arXiv preprint arXiv:1811.05868}, 2018.

\bibitem[Shi et~al.(2023)Shi, Liang, and Wang]{shi2023lmc}
Zhihao Shi, Xize Liang, and Jie Wang.
\newblock Lmc: Fast training of gnns via subgraph sampling with provable convergence.
\newblock \emph{arXiv preprint arXiv:2302.00924}, 2023.

\bibitem[Velickovic et~al.(2017)Velickovic, Cucurull, Casanova, Romero, Lio, Bengio, et~al.]{velickovic2017graph}
Petar Velickovic, Guillem Cucurull, Arantxa Casanova, Adriana Romero, Pietro Lio, Yoshua Bengio, et~al.
\newblock Graph attention networks.
\newblock \emph{stat}, 1050\penalty0 (20):\penalty0 10--48550, 2017.

\bibitem[Vieira(2014)]{vieira2014gumbel}
Tim Vieira.
\newblock Gumbel-max trick and weighted reservoir sampling, 2014.
\newblock URL \url{http://timvieira.github.io/blog/post/2014/08/01/gumbel-max-trick-and-weighted-reservoir-sampling/}.

\bibitem[Wang et~al.(2021)Wang, Liu, Fan, Sun, and Yu]{wang2021dskreg}
Yu~Wang, Zhiwei Liu, Ziwei Fan, Lichao Sun, and Philip~S Yu.
\newblock Dskreg: Differentiable sampling on knowledge graph for recommendation with relational gnn.
\newblock In \emph{Proceedings of the 30th ACM International Conference on Information \& Knowledge Management}, pp.\  3513--3517, 2021.

\bibitem[Williams(1992)]{williamsSimpleStatisticalGradientfollowing1992}
Ronald~J. Williams.
\newblock Simple statistical gradient-following algorithms for connectionist reinforcement learning.
\newblock \emph{Machine Learning}, 1992.
\newblock ISSN 0885-6125.
\newblock \doi{10.1007/bf00992696}.

\bibitem[Wu et~al.(2022)Wu, Sun, Zhang, Xie, and Cui]{wu2022graph}
Shiwen Wu, Fei Sun, Wentao Zhang, Xu~Xie, and Bin Cui.
\newblock Graph neural networks in recommender systems: a survey.
\newblock \emph{ACM Computing Surveys}, 55\penalty0 (5):\penalty0 1--37, 2022.

\bibitem[Yang et~al.(2016)Yang, Cohen, and Salakhudinov]{yang2016revisiting}
Zhilin Yang, William Cohen, and Ruslan Salakhudinov.
\newblock Revisiting semi-supervised learning with graph embeddings.
\newblock In \emph{International conference on machine learning}, pp.\  40--48. PMLR, 2016.

\bibitem[Yoon et~al.(2021)Yoon, Gervet, Shi, Niu, He, and Yang]{PASS}
Minji Yoon, Th{\'e}ophile Gervet, Baoxu Shi, Sufeng Niu, Qi~He, and Jaewon Yang.
\newblock Performance-adaptive sampling strategy towards fast and accurate graph neural networks.
\newblock In \emph{Proceedings of the 27th ACM SIGKDD Conference on Knowledge Discovery \& Data Mining}, pp.\  2046--2056, 2021.

\bibitem[You et~al.(2022)You, Lu, Zhou, Fu, and Lin]{you2022early}
Haoran You, Zhihan Lu, Zijian Zhou, Yonggan Fu, and Yingyan Lin.
\newblock Early-bird gcns: Graph-network co-optimization towards more efficient gcn training and inference via drawing early-bird lottery tickets.
\newblock In \emph{Proceedings of the AAAI Conference on Artificial Intelligence}, volume 36(8), pp.\  8910--8918, 2022.

\bibitem[Yu et~al.(2022)Yu, Wang, Wang, Liu, Yang, and Ji]{yu2022graphfm}
Haiyang Yu, Limei Wang, Bokun Wang, Meng Liu, Tianbao Yang, and Shuiwang Ji.
\newblock Graphfm: Improving large-scale gnn training via feature momentum.
\newblock In \emph{International Conference on Machine Learning}, pp.\  25684--25701. PMLR, 2022.

\bibitem[Yun et~al.(2019)Yun, Jeong, Kim, Kang, and Kim]{yun2019graph}
Seongjun Yun, Minbyul Jeong, Raehyun Kim, Jaewoo Kang, and Hyunwoo~J Kim.
\newblock Graph transformer networks.
\newblock \emph{Advances in neural information processing systems}, 32, 2019.

\bibitem[Zeng et~al.(2019)Zeng, Zhou, Srivastava, Kannan, and Prasanna]{graphsaint}
Hanqing Zeng, Hongkuan Zhou, Ajitesh Srivastava, Rajgopal Kannan, and Viktor Prasanna.
\newblock {GraphSAINT}: Graph sampling based inductive learning method.
\newblock \emph{arXiv preprint arXiv:1907.04931}, 2019.

\bibitem[Zhang et~al.(2024)Zhang, Wang, Huang, Yue, Wang, Zimmermann, Zhou, Cheng, Zeng, and Liang]{zhang2024lottery}
Guibin Zhang, Kun Wang, Wei Huang, Yanwei Yue, Yang Wang, Roger Zimmermann, Aojun Zhou, Dawei Cheng, Jin Zeng, and Yuxuan Liang.
\newblock Graph lottery ticket automated.
\newblock In \emph{The Twelfth International Conference on Learning Representations}, 2024.

\bibitem[Zhao et~al.(2023)Zhao, Dong, Hanjalic, and Khosla]{zhao2023multi}
Tianqi Zhao, Ngan~Thi Dong, Alan Hanjalic, and Megha Khosla.
\newblock Multi-label node classification on graph-structured data.
\newblock \emph{Trans. Mach. Learn. Res.}, 2023, 2023.

\bibitem[Zheng et~al.(2022)Zheng, Wang, Liu, Li, Zhang, Jin, Yu, and Pan]{zheng2022graph}
Xin Zheng, Yi~Wang, Yixin Liu, Ming Li, Miao Zhang, Di~Jin, Philip~S Yu, and Shirui Pan.
\newblock Graph neural networks for graphs with heterophily: A survey.
\newblock \emph{arXiv preprint arXiv:2202.07082}, 2022.

\bibitem[Zhu et~al.(2020)Zhu, Yan, Zhao, Heimann, Akoglu, and Koutra]{zhu2020beyond}
Jiong Zhu, Yujun Yan, Lingxiao Zhao, Mark Heimann, Leman Akoglu, and Danai Koutra.
\newblock Beyond homophily in graph neural networks: Current limitations and effective designs.
\newblock \emph{Advances in neural information processing systems}, 33:\penalty0 7793--7804, 2020.

\bibitem[Zou et~al.(2019)Zou, Hu, Wang, Jiang, Sun, and Gu]{LADIES}
Difan Zou, Ziniu Hu, Yewen Wang, Song Jiang, Yizhou Sun, and Quanquan Gu.
\newblock Layer-dependent importance sampling for training deep and large graph convolutional networks.
\newblock In \emph{Advances in Neural Information Processing Systems 32: Annual Conference on Neural Information Processing Systems 2019, NeurIPS 2019}, pp.\  11247--11256, 2019.

\end{thebibliography}
\bibliographystyle{tmlr}

\appendix
\section{Off-Policy Sampling Setup}
\label{appendix:off-policy}
In this Appendix, we discuss the technical and mathematical challenges around our setup that resulted in our off-policy learning setup. In each layer $l$ of the GFlowNet, we aim to sample exactly $k$ out of $n$ nodes. An initially natural setup would be to use the distribution over $k$-subsets of $\mathcal{N}(K^{(l-1)})$ \citep{ahmed2023simple}. Using Bayes theorem,
\begin{equation}
    q(\mathcal{V}^{(l)}|\mathcal{V}^{(0)}, \ldots, \mathcal{V}^{(l-1)}, k)= \frac{I[|\mathcal{V}^{(l)}|=k] q(\mathcal{V}^{(l)}|\mathcal{V}^{(0)}, \ldots, \mathcal{V}^{(l-1)})}{\sum_{\mathcal{V}^{'(l+1)}} I[|\mathcal{V}^{'(l+1)}|=k] q(\mathcal{V}^{(l)}|\mathcal{V}^{(0)}, \ldots, \mathcal{V}^{(l-1)})}.
\end{equation}
When conditioned on $k$, $q$ assigns 0 probability to sets of nodes $\mathcal{V}^{(l)}$ that do not sample exactly $k$ new nodes (that is, when $|\mathcal{V}^ {(l)}|\neq k$). However, this requires renormalizing the distribution, which is the function of the denominator term on the right-hand side. Note that this sum is over an exponential number of elements, namely $2^{|\mathcal{N}(K^{(l-1)})|}$, and naive computation is clearly intractable. SIMPLE \citep{ahmed2023simple} provides an optimized dynamic programming algorithm for computing this normalization constant. However, it scales polynomially in $|\mathcal{N}(K^{(l-1)})|$ and $k$, and in our experiments, computing the normalizer is a bottleneck already for mid-sized graphs like Reddit. 

Therefore, we decided to circumvent having to compute $q(\mathcal{V}^{(l)}|\mathcal{V}^{(0)}, \ldots, \mathcal{V}^{(l-1)}, k)$ by sampling using the Gumbel-Top-k trick (Equation \ref{gumbel}) to ensure we always add exactly $k$ nodes. However, we are now in an off-policy setting: The samples using Equation \ref{gumbel} are distributed by $q(\mathcal{V}^{(l)}|\mathcal{V}^{(0)}, \ldots, \mathcal{V}^{(l-1)}, k)$, not by $q(\mathcal{V}^{(l)}|\mathcal{V}^{(0)}, \ldots, \mathcal{V}^{(l-1)})$, and so we sample from a different distribution than the one we use to compute the loss. Previous work \citep{malkin2022gflownets-off} showed that the Trajectory Balance loss is amenable to off-policy training without importance sampling and weighting without introducing high variance. This is important since importance weighting would require us to weight by $q(\mathcal{V}^{(l)}|\mathcal{V}^{(0)}, \ldots, \mathcal{V}^{(l-1)}) / q(\mathcal{V}^{(l)}|\mathcal{V}^{(0)}, \ldots, \mathcal{V}^{(l-1)}, k)$, reintroducing the need to compute $q(\mathcal{V}^{(l)}|\mathcal{V}^{(0)}, \ldots, \mathcal{V}^{(l-1)}, k)$. 

The off-policy benefits of the Trajectory Balance loss provide a strong argument over more common Reinforcement Learning setups. Off-policy training in Reinforcement Learning usually requires importance weighting to be stable, which is not tractable in our setting.

\section{Experimental Details}
\label{appendix-experimental}
For all experiments, we used as architecture the Graph Convolutional Network~\citep{kipf2016semi}, with two layers, a hidden size of 256, a batch size of 256, and a sampling size of 256 nodes per layer. We implemented the GCNs in \myMethod{} via PyTorch Geometric \citep{Fey/Lenssen/2019}. We train for 50 epochs on Cora, Citeseer, and Reddit; 100 epochs on BlogCat, DBLP, Flickr, ogbn-products, Pubmed, snap-patents, and Yelp; and 150 epochs on ogbn-arxiv and ogbn-proteins. The ogbn-proteins and BlogCat datasets do not contain node features, and instead we learn node embeddings for them of dimension 128 for ogbn-proteins, and 64 for BlogCat.

Our experiments were carried out in a single-node cluster setup. We conducted our experiments on a machine with Nvidia RTX A4000 GPU (16GB GPU memory), Nvidia A100 (40GB GPU memory), and Nvidia RTX A6000 GPU (48GB GPU memory) and each machine had 48 CPUs. In total, we estimate that our experiments took 200 compute days. 

\subsection{Hyperparameter Tuning}
We tune the hyperparameters of \myMethod{} using a random search strategy with the goal of maximizing the accuracy of the validation dataset. We used Weights and Biases for hyperparameter tuning \footnote{ \url{https://wandb.ai} }. The best-performing hyperparameters for every dataset can be found in our repository \url{https://anonymous.4open.science/r/GRAPES}. The following are the hyperparameters that we tuned: the learning rate of the GFlowNet, the learning rate of the classification GCN, and the scaling parameter $\alpha$. We used the log uniform distribution to sample the aforementioned hyperparameters with the values from the following ranges, respectively, $[1e-6, 1e-2]$, $[1e-6, 1e-2]$, and [1e2, 1e6]. We kept the other hyperparameters, such as the batch size and hidden dimension of the GCN. We used the Adam optimizer \citep{kingma2014adam} for $\operatorname{GCN}_{\operatorname{C}}$ and $\operatorname{GCN}_{\operatorname{S}}$. 
%The number of epochs was selected between 50, 100, and 150 depending on the performance on the validation set. 

We did a hyperparameter sensitivity analysis performed using Weights \& Biases \cite{wandb} for training of GRAPES-GFN on Yelp. Importance measures how strongly each parameter influences the validation accuracy (higher = more critical), while correlation shows the direction and magnitude of the linear relationship (positive = increasing the parameter tends to increase the accuracy, negative = the opposite). From the table below, $lr_{gc}$, the learning rate for the classification GCN, has the strongest effect on accuracy and is negatively correlated, suggesting that higher learning rates can harm performance. Meanwhile, $lr_{gf}$ and $\alpha$ show smaller impacts and weaker correlations, indicating that the model is less affected by changes in these two parameters.
\begin{table}[h]
    \centering
    \caption{Importance and correlation of configuration parameters for training GRAPES-GFN on Yelp.}
    \begin{tabular}{lcc}
        \toprule
        \textbf{Config Parameter} & \textbf{Importance} & \textbf{Correlation} \\
        \midrule
        lr\_gc & 0.633 & -0.610 \\
        lr\_gf  GCN) & 0.280 & 0.055 \\
        $\alpha$  & 0.087 & 0.097 \\
        \bottomrule
    \end{tabular}
    
    \label{tab:config_params}
\end{table}

\subsection{Baselines}

\begin{table}[]
    \caption{Differences in experimental setups in related work, obtained from the original publications and their official implementations. *This indicates the total budget in terms of nodes sampled across all layers.}
    \label{tab:diff-experiment-setup}
    \centering
    \resizebox{\textwidth}{!}{
    \begin{tabular}{lccccc}
    \toprule
         & \bf FastGCN~\citep{fastGCN} & \bf LADIES~\citep{LADIES} & \bf GraphSAINT~\citep{graphsaint} & \bf AS-GCN~\citep{AS-GCN} & \bf GAS~\citep{fey2021gnnautoscale} \\
         \midrule
        Architecture & GCN & GCN & GCN & GCN+attention & GCN,GCNII,GAT,GIN,APPNP,PNA\\
        Hidden size & (16, 128) & 256 & (128, 256, 512, 2048) & (16,256)  & (256, 512, 1024, 2048) \\
        Number of layers & 2 & 5 & (2, 4, 5) & 2 & (2, 4, 64) \\
        Batch size & (256, 1024) & 512 & (400, 512, 1000) & 256 & (1,2,5,40,12,100) \\
        Nodes per layer & (100, 400) & (5, 64, 512) & (4500, 6000, 8000)* & (128, 256, 512) & No sampling\\
        Training epochs & (100, 200, 300) & 300 & 2000 & (50,100,300) & (300, 400, 500, 1000) \\
    \bottomrule
    \end{tabular}
    }

\end{table}

For a fair comparison, we adjusted the implementations of the baselines so that the only difference is the sampling methods and the rest of the training conditions are kept the same. In the following, we explain the details of the modifications to each of the baselines.

For LADIES, we used the official implementation, which also contains an implementation of FastGCN. We changed the nonlinear activation function from ELU to ReLU, and we removed any linear layers after the two layers of the GCN, set dropout to zero, and disabled early stopping. We also noticed that the original LADIES implementation divided the target nodes into mini-batches, not from the entire graphs as we do, but into random fragments. This means that LADIES and FastGCN do not see all the target nodes in the training data. We kept this setting unchanged because otherwise it would have significantly slowed down the training of these two methods.

For GraphSAINT, we noticed that the GNN consists of two layers of higher-order aggregators, which are a combination of GraphSage-mean \citep{hamilton2017inductive} and MixHop \citep{abu2019mixhop}, and a linear classification layer at the end. Moreover, the original implementation of GraphSAINT is only applicable to inductive learning on the graphs, where the training graph only contains the training nodes and is entirely different from the validation and test graphs, where only the nodes from the validation set and test are available, respectively. We argue that in transductive learning, unlike inductive learning, the motivation to scale to larger graphs is higher since the validation and test nodes are also available during training, and therefore, the processed graph is larger. 
%We noticed that the original GraphSAINT implementation is less applicable to transductive learning because the sampling step became significantly slower after modifying the implementation to adjust to the transductive graphs.
Finally, to keep all the configurations the same as \myMethod{}, we used PyTorch Geometric's function for GraphSAINT Random Walk sampler with depth two, which showed the best performance across the three variations of GraphSAINT. Therefore, we use the same GCN and data loader (transductive) as ours and use GraphSAINT to sample a subgraph of nodes for training. We use the node sampler setting since it is the only setting that allows specifying different sampling budgets, and therefore, can be compared to layer-wise methods with the same sampling budget. We also removed the early stopping and used the same number of epochs as \myMethod{}. Note that for Figure \ref{fig:sampling-effect}, we need to change the sample size while the Random Walk sampler can only sample equal to the batch size per layer. Therefore, for these experiments, we used the Node Sampler version of GraphSAINT, which allows us to change the sample size. For instance, for sample size $32$ per layer, we set GraphSAINT's Node Sampler's sample size to $256+32+32=320$ to sample a subgraph with 320 nodes.
Please refer to our repository for more details about the implementation of GraphSAINT.

For GAS, we used the original implementation. However, we changed the configuration of the GCN to have a two-layer GCN with 256 hidden units. We turned off dropout, batch normalization and residual connections in the GCN. We also removed early stopping for the training.

For AS-GCN we removed the attention mechanism used in the GCN classifier. Their method also uses attention in the sampler, which is separate from the classifier, so we keep it.

For PASS, we changed the inference to full-batch to match the rest of the baselines. We set the parameter \texttt{sampling\_scope} to $512$, because in the original experiments of the paper, this value was set to twice the batch size. For Flickr, ogbn-arxiv, and ogbn-proteins, this value results in OOM; therefore, we set it to $128$, $128$, and $256$, respectively. For the datasets that we report OOM in tables \ref{tab1} and \ref{tab2}, we tested all possible \texttt{sampling\_scope} between $512$ and $16$.

\section{Dataset statistics}
\label{app:data_statistics}
We present the statistics of the datasets used in our experiments in Table~\ref{tab:data_statistics}. The splits that we used for Cora, Citeseer, and Pubmed correspond to the ``full'' splits, in which the label rate is higher than in the ``public'' splits. For BlogCat, we take the average accuracy of all the methods across the three available splits provided by \citep{zhao2023multi}. For DBLP and snap-patents, we use the average of ten random splits because these two datasets had no predefined splits. 

\begin{table}[t]
\caption{Statistics of the datasets used in our experiments. The label rate indicates the percentage of nodes used for training. ogbn-proteins and BlogCat do not contain node features, and instead, we learn embeddings for nodes in these datasets.}
\label{tab:data_statistics}
\centering
\begin{tabular}{lcccccc}
\toprule
\bf Dataset   & \bf Task    & \bf Nodes & \bf Edges  & \bf Features & \bf Classes & \bf Label Rate (\%) \\
\midrule
Cora          & multi-class & 2,708     & 5,278      & 1,433        & 7           & 44.61               \\
CiteSeer      & multi-class & 3,327     & 4,552      & 3,703        & 6           & 54.91               \\
PubMed        & multi-class & 19,717    & 44,324     & 500          & 3           & 92.39               \\
Reddit        & multi-class & 232,965   & 11,606,919 & 602          & 41          & 65.86               \\
DBLP          & multi-label & 28,702 & 136,670 & 300 & 4 & 80.00 \\
Flickr        & multi-class & 89,250    & 449,878    & 500          & 7           & 50.00               \\
snap-patents        & multi-class &   2,923,922  &  27,945,092   &  269        &    5        & 50.00               \\
Yelp          & multi-label & 716,847   & 6,977,409  & 300          & 100         & 75.00               \\
ogbn-proteins        & multi-label &   132,534 &  79,122,504   &  ---        &    112        & 65.30               \\
BlogCat      & multi-label &   10,312 &  667,966   &  ---        &    39        & 60.00               \\
ogbn-arxiv    & multi-class & 169,343   & 1,157,799  & 128          & 40          & 53.70               \\
ogbn-products & multi-class & 2,449,029 & 61,859,076 & 100          & 47          & 8.03                \\
\bottomrule
\end{tabular}
\end{table}

\section{Entropy as Node preference measure}
\label{appendix-entropy}
\begin{figure}[t]
    \centering
    \includegraphics[width=\textwidth]{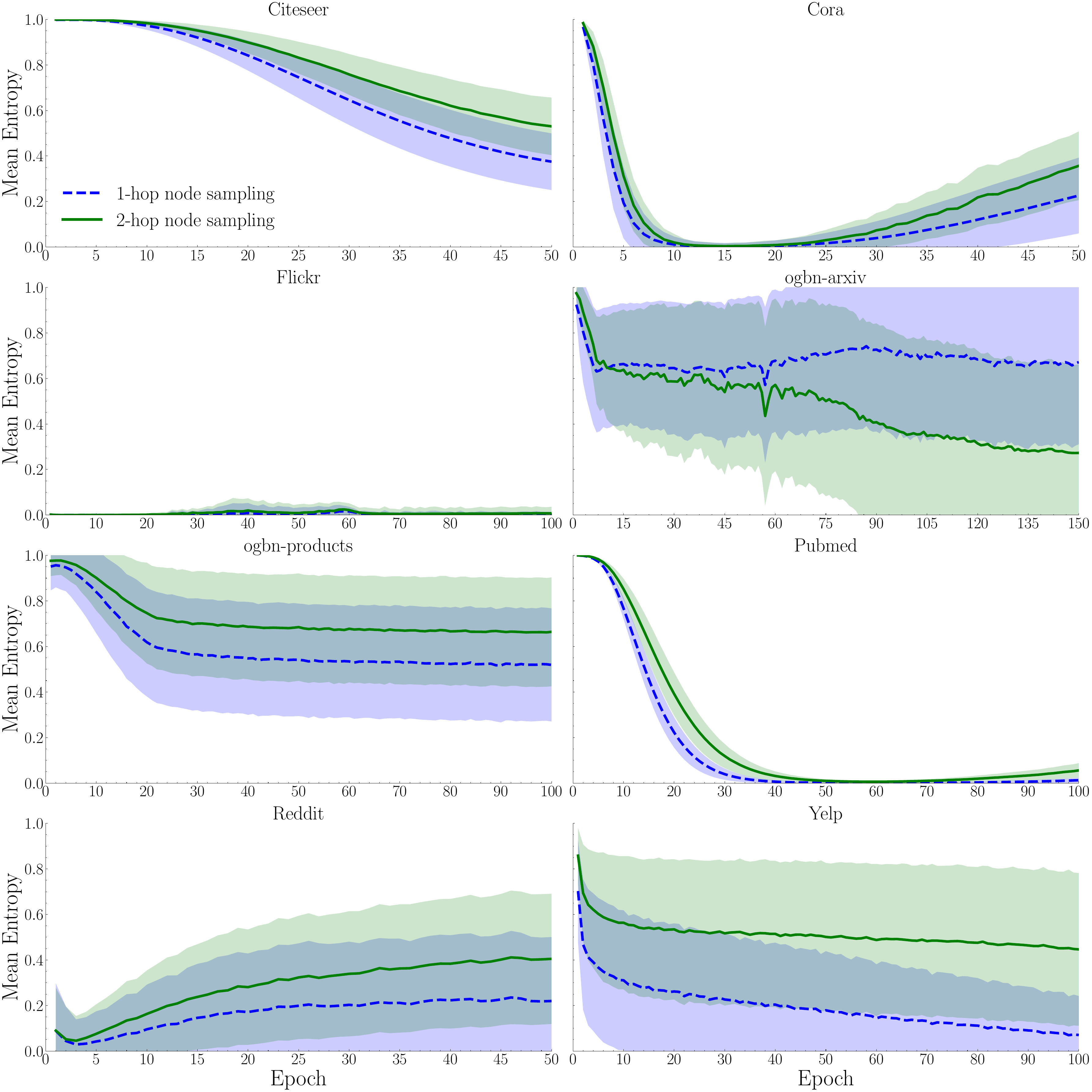}
    \caption{Combined entropy plots for Citeseer, Cora, Flickr, ogbn-arxiv, ogbn-products, Pubmed, Reddit, and Yelp showcasing the mean entropy. The shaded region indicates the standard deviation of the entropy \emph{across nodes}. The plots compare 1-hop node sampling against 2-hop node sampling.}
    \label{fig:combined_entropy_plot1}
\end{figure}

\begin{figure}[t]
    \centering
    \includegraphics[width=\textwidth]{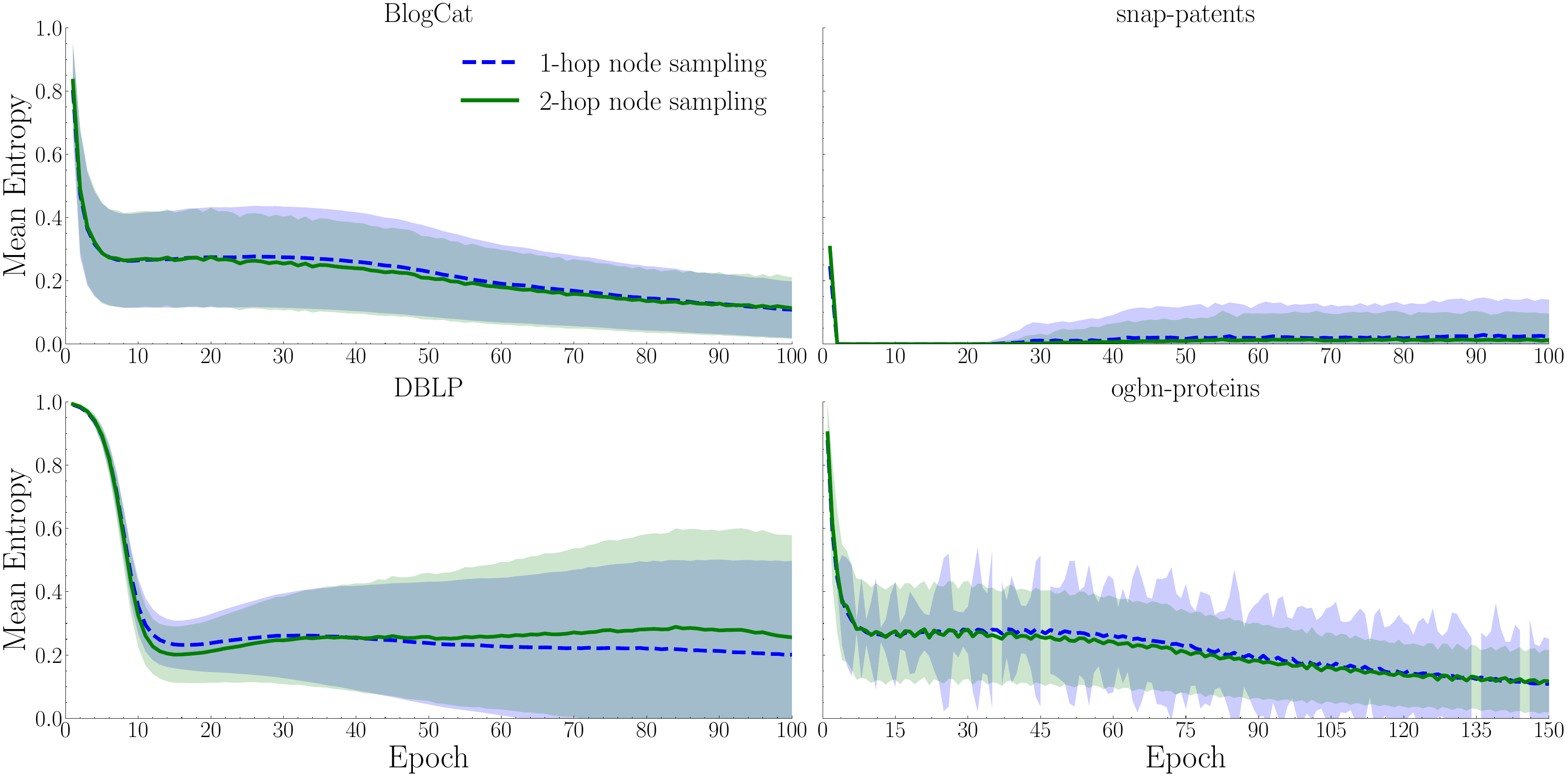}
    \caption{Combined entropy plots for BlogCat, snap-patents, DBLP, and ogbn-proteins, showcasing the mean entropy across epochs.  The shaded region indicates the standard deviation of the entropy \emph{across nodes}. The plots compare 1-hop node sampling against 2-hop node sampling.}
    \label{fig:combined_entropy_plot2}
\end{figure}

Figures \ref{fig:combined_entropy_plot1} and  \ref{fig:combined_entropy_plot2} show the mean and standard deviation of entropy in base two of all the datasets. We calculate the mean entropy as the following:

\begin{align}
    E = \frac{1}{n}\sum_{i=1}^n p_i \cdot \log_2(p_i) + (1-p_i)\cdot \log_2(1-p_i) 
\end{align}

where $n$ is the number of neighbors of the nodes sampled in the previous layer and $p_i$ is the probability of inclusion for each node, which is the output of the GFlowNet. As the figures show, for certain small datasets (Cora, Citeseer, Pubmed, Flickr) the mean entropy is: 1) very close to 1, indicating that \myMethod{} prefers every nodes with the probability close to $0.5$, or 2) close to 0 but also with a low standard deviation, meaning that it equally prefers the majority of the nodes with the probability 1 or 0. On the contrary, for the large datasets (Reddit, Yelp, ogbn-arxiv, ogbn-products) by the end of training, the average entropy is lower than 1, with a standard deviation around $0.3$ indicating that \myMethod{} learns different preferences over different nodes, some with a probability close to 1, and some close to 0.

We analyzed the label distribution of the sampled subgraphs via GRAPES-GF for Yelp, by counting the percentage of nodes having each label and calculating the difference between the original and sampled graphs, i.e., percentage of nodes having label $c$ in the original graph minus percentage of nodes having label $c$ in the sampled graph. As shown in the figure, all the values are positive, indicating that a higher ratio of nodes have each label in the original graph. Moreover, this difference increases at the end of training. This means that among the 100 labels in Yelp, GRAPES prefers the nodes with fewer labels or that are even single-labeled. This may make the classification task easier and result in a lower loss. Moreover, among the 100 classes, some are sampled the most and some the least. Figure \ref{fig:distribution_plot} below shows this trend.
\begin{figure}[h]
    \centering
    \includegraphics[width=\textwidth]{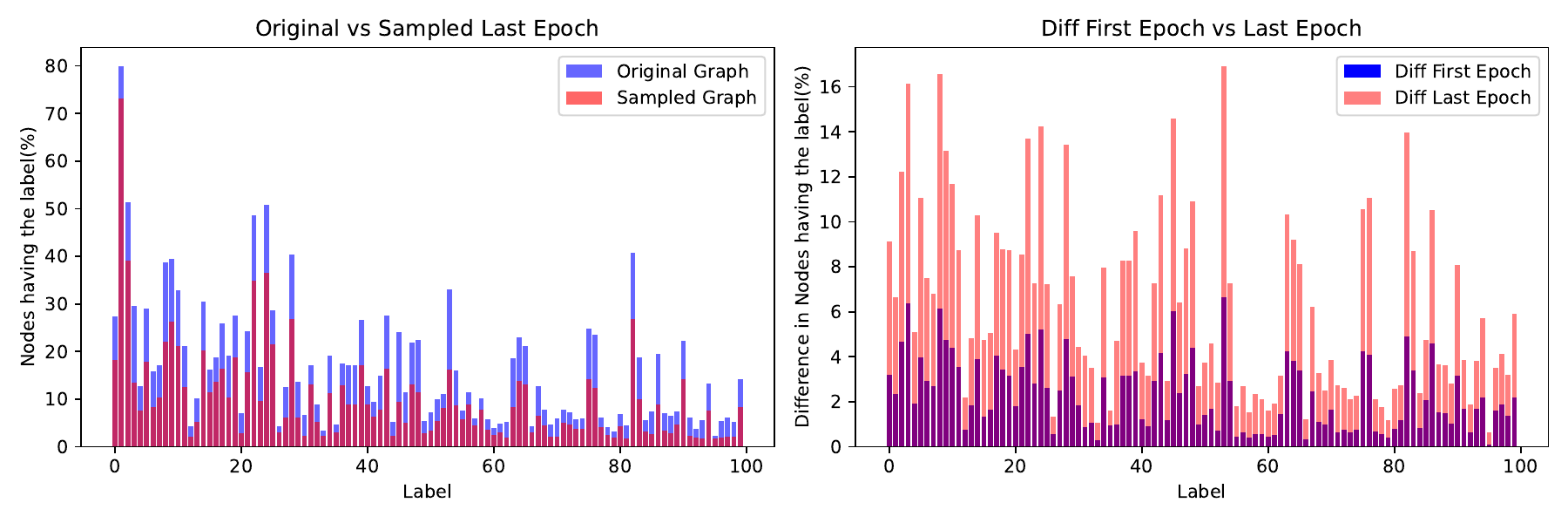}
    \caption{Comparison of Label distributions between GRSAPES-GFN and the original graph for Yelp. The left figure shows the percentage of nodes in the last epoch of both graphs that have each label. The right shows the difference between those percentages in the original graph and the sampled graph at the beginning and end of training. }
    \label{fig:distribution_plot}
\end{figure}

\section{GPU Memory usage comparison between \myMethod{} and GAS}

We compared different variants of \myMethod{}, with different sample sizes (32, 256), with GAS \citep{fey2021gnnautoscale}, which is a non-sampling method. Figure \ref{fig:memory-gas-grapes} shows the GPU memory allocation (MB), on a logarithmic scale for GRAPES-32, GRAPES-256, and GAS. The three graph methods exhibit distinct performance characteristics across various datasets. We used the \texttt{max\_memory\_allocated} function in PyTorch to measure the GPU memory allocation.\footnote{\url{https://pytorch.org/docs/stable/generated/torch.cuda.max_memory_allocated.html}} Since this function measures the maximum memory allocation since the beginning of the program, where the memory measurement is done is not relevant.

\section{GFlowNet details}

This section provides additional details on GFlowNets in general and the GFlowNet version of GRAPES.

\subsection{Generative Flow Networks} 
Generative Flow Networks (GFlowNets) \citep{bengio2021flow, bengio2021gflow-fundation} are generative models that can sample from a very large structured space. GFlowNets construct the structure in multiple generation steps. Compared to Reinforcement Learning approaches, GFlowNets learn to sample \emph{in proportion} to a given reward function, while in Reinforcement Learning, reward functions are maximized. This feature of GFlowNets encourages sampling diverse sets of high-quality structures, instead of only considering the single best structure. GFlowNets have been used in several applications like molecule design and material science \citep{bengio2021flow, gao2022molecule, jain2022biological}, Bayesian structure learning \citep{deleu2022bayesian}, scientific discovery \citep{jain2023scientific}, and GNN explainability \citep{li2023dag}. Similar to our work, the latter utilizes a GFlowNet to sample subgraphs. However, this method \emph{explains} a trained GNN and is not used to scale GNN training to large graphs.

\subsection{GFlowNet and Trajectory Balance Loss}
We first give a brief overview of GFlowNets \citep{bengio2021flow,bengio2021gflow-fundation} and the trajectory balance loss \citep{malkin2022trajectorybalance}. 
Let $\mathcal{G}_F=(\mathcal{S}, \mathcal{A}, \mathcal{S}_0, \mathcal{S}_f, R)$ denote a GFlowNet learning problem. Here, $\mathcal{S}$ is a finite set of states that forms a directed graph with $\mathcal{A}$, a set of directed edges representing actions or transitions between states. $\mathcal{S}_0 \subset \mathcal{S}$ is the set of initial states, $\mathcal{S}_f\subset \mathcal{S}$ is the set of terminating states,\footnotemark~and $R: \mathcal{S}_f\rightarrow \mathbb{R}_+$ is the reward function defined on terminating states. At time $t$, a particular $a_t \in \mathcal{A}$ indicates the action taken to transition from state $s^{(t-1)}$ to $s^{(t)}$. A trajectory $\tau$ is a path through the graph from an initial state $s^{(0)}$ to a terminating state $s^{(n)}\in\mathcal{S}_f$: $\tau=(s^{(0)}\to\ldots\to s^{(n)})$. A GFlowNet is a neural network that learns to transition from an initial state $s^{(0)}$ to a terminating state where the reward $R(s^{(n)})$ is given. The goal of the GFlowNet is to ensure that following the forward transition probabilities $P_F(s^{(t)}|s^{(t-1)})$ leads to final states $s \in \mathcal{S}_f$ with probability in proportion to the reward $R$ \citep{bengio2021flow}. %`Flow' refers to the unnormalized probabilities of transitioning between the states \citep{bengio2021gflow-fundation}. 
The \emph{Trajectory Balance (TB)} loss \citep{malkin2022trajectorybalance} is developed with this goal. For a trajectory $\tau=(s^{(0)} \to \ldots \to s^{(n)})$, the TB loss is: 
\begin{equation}
\label{trajectory balance}
    \mathcal{L}_{TB}(\tau) = \left (\log \frac{Z(s^{(0)}) \prod_{t=1}^n P_F(s^{(t)}|s^{(t-1)})}{R(s^{(n)})\prod_{t=1}^n P_B(s^{(t-1)}|s^{(t)})} \right)^2,
\end{equation}
where $Z:\mathcal{S}_0\rightarrow \mathbb{R}_+$ computes the total flow of the network from the starting state $s^{(0)}$ and $P_F$ and $P_B$ are the forward and the backward transition probabilities between the states, where both can be parameterized by a neural network \citep{malkin2022trajectorybalance}. 

\footnotetext{Technically, GFlowNets have unique source and terminal (or `sink') states $s_s$ and $s_f$. The source state has an edge to all initial states, and all terminating states have an edge to the terminal state.}

\subsection{GFlowNet Design: States, Actions, and Reward}
\label{sec:gflownet-design}
Next, we explain our choice of $\mathcal{G}_F$, that is, the states, actions, terminating states, and reward function, and the coupling of our GFlowNet with the sampling policy $q$. A state $s^{(l-1)}\in \mathcal{S}$ represents a sequence of sets of nodes $s^{(l-1)}=(\mathcal{V}^{(0)}, \ldots, \mathcal{V}^{(l-1)})$ sampled so far. An action from $s^{(l-1)}$ represents choosing $k$ nodes without replacement among $\mathcal{N}(K^{(l-1)})$. This forms the set of nodes $\mathcal{V}^{(l)}$ in the next layer. Therefore, in an $L$ layer GCN, we construct a sequence of $L$ sets of nodes to reach a terminating state. 
%Thus, in our design, obtaining the adjacency matrix of the next layer is equivalent to transitioning to a new state. 
%Figure \ref{fig1} shows our DAG structure: the trajectories will have $L$ actions and $L+1$ states (including the initial state). 

We define the optimal sampling policy as having the lowest classification loss in expectation. Therefore, a set of $k$ nodes with a lower classification loss than another set must have a higher probability. Our goal is to design $\mathcal{G}_F$ so that it learns the forward and backward transition probabilities proportional to a given reward. We define the reward function as below:
\begin{equation}
    R(s^{(L)}) = R(\mathcal{V}^{(0)}, \ldots, \mathcal{V}^{(L)}) := \exp(-\alpha\cdot \mathcal{L}_{\operatorname{C}}(X, Y, K^{(0)}, \ldots, K^{(L)})),
\end{equation}
where $\mathcal{L}_{\operatorname{C}}$ is the classification loss and $\alpha$ is a \emph{scaling parameter}, which we explain in Section \ref{sec:rewardscaling}. %This reward function prefers adjacency matrices $A_0$ to $A_L$ that lead to a lower loss by $\operatorname{GCN}_{\operatorname{S}}$. 
%For simplicity, we denote each state $s^{(t)}$ as its corresponding adjacency matrix $A_t$. 

\textbf{Forward Probability}.
The forward probability $P_F(s^{(l-1)}|s^{(l)})$ in GRAPES is the sampling policy $q(\mathcal{V}^{(l)}|\mathcal{V}^{(0)}, \ldots, \mathcal{V}^{(l-1)})$ defined in Section \ref{sec:sampling-policy}. 

\textbf{Backward Probability}.
Trajectory balance (Equation \ref{trajectory balance}) also requires defining the probability of transitioning backwards through the states. The backward probability is a distribution over all parents of a state.
This distribution is not required in our setup, as the state representation $s^{(l)}=(\mathcal{V}^{(0)}, \ldots, \mathcal{V}^{(l)})$ saves the trajectory taken through $\mathcal{G}_F$ to get to $s^{(l)}$. This means the graph for the GFlowNet learning problem $\mathcal{G}_F$ is a tree, as each state $s^{(l)}=(\mathcal{V}^{(0)}, \ldots, \mathcal{V}^{(l-1)}, \mathcal{V}^{(l)})$ has exactly 1 parent, namely $s^{(l-1)}=(\mathcal{V}^{(0)}, \ldots, \mathcal{V}^{(l-1)})$. Since each state has a single parent, we find that $P_B(s^{(l-1)}|s^{(l)})=1$ when we retrace the trajectory. We pass the information on when a node is added to the GFlowNet by adding an identifier to the nodes' embeddings that indicates in what layer it was sampled, as explained in Section \ref{sec:sampling-policy}.

\textbf{Loss derivation} Combining our setup with the trajectory balance loss (Equation \ref{trajectory balance}), the \myMethod{} loss is
\begin{align}
\label{trajectoryshort}
    \mathcal{L}_{\operatorname{GFN}}(X,Y, \mathcal{V}^ {(0)})= \Big(&\log \frac{Z(\mathcal{V}^{(0)})\prod_{l=1}^L P_F(s^{(l)}|s^{(l-1)})}{R(s^{(l)})} \Big)^2 \\
    = \Big(&\log Z(\mathcal{V}^{(0)}) + \sum_{l=1}^L \log q(\mathcal{V}^{(l)}|\mathcal{V}^{(0)}, \ldots, \mathcal{V}^{(l-1)}) + \\
    &\alpha \cdot \mathcal{L}_{\operatorname{C}}(X, Y, K^{(0)}, \ldots, K^{(L)})\Big)^2\notag
\end{align}
We model the initial-state-dependent normalizer $Z(s^{(0)})$ in Equation \ref{trajectory balance} with a trainable GCN, namely $\operatorname{GCN}_Z(\mathcal{V}^{(0)})$. It predicts the normalizer conditioned on the target nodes given. It is trained together with $\operatorname{GCN}_{\operatorname{S}}$ by minimizing the trajectory balance loss. 

We note that, like in \myMethod{}-RL, we use off-policy sampling from $q(\mathcal{V}^{(l)}|\mathcal{V}^{(0)}, \ldots, \mathcal{V}^{(l-1)}, k)$ to train the GFlowNet. See Appendix \ref{appendix:off-policy} for additional details. 

\subsection{Reward scaling}
\label{sec:rewardscaling}
In our experiments, we noticed that with the bigger datasets, the GFlowNet is more affected by the log-probabilities than the reward from the classification $\operatorname{GCN}_{\operatorname{C}}$. The reason that the term $\log q(\mathcal{V}^{(l)}|\mathcal{V}^{(0)}, \ldots, \mathcal{V}^{(l-1)})$ dominates $\mathcal{L}_{\operatorname{GFN}}$ is that $\log q(\mathcal{V}^{(l)}|\mathcal{V}^{(0)}, \ldots, \mathcal{V}^{(l-1)})=\sum_{i\in \mathcal{N}(K^{(l-1)})}\log p_i $, which sums over $|\mathcal{N}(K^{(l-1)})|$ elements. Given a batch size of $256$, this neighborhood can be as big as $52,000$ nodes, resulting in summing $52,000$ log-probabilities. The majority of the probabilities $p_i$ are values close to zero. Therefore, the above sum would be a large negative number. Since the loss $\mathcal{L}_{\operatorname{C}}$ is, in our experiments, often quite close to zero, the log-probability and its variance dominate the loss. Therefore, we add the hyperparameter $\alpha$ to the reward and tune it in our experiments. 

\section{\myMethod{} and the straight-through estimator}
\label{sec:app-sthrough}

\begin{table}[t]
    \centering
    \caption{Comparison between \myMethod{} and variants using the straight-through estimator (STE).}
    \label{tab:ste-comparison}
    \begin{tabular}{ccccc}
    \toprule
         & Cora & Flickr & ogbn-arxiv & Yelp \\
    \midrule
         Best \myMethod{} & 87.62 $\pm$ 0.48 & 49.54 $\pm$ 0.67 & 62.58 $\pm$ 0.64 & 44.57 $\pm$ 0.88 \\
         STE              & 87.95 $\pm$ 0.18 & 46.33 $\pm$ 1.39 & 61.95 $\pm$ 0.32 & 37.29 $\pm$ 0.15 \\
         STE$_\text{N}$   & 87.03 $\pm$ 0.35 & 46.29 $\pm$ 1.47 & 60.13 $\pm$ 0.41 & 15.63 $\pm$ 0.10 \\
    \bottomrule
    \end{tabular}
\end{table}

The straight-through estimator~\cite{bengio2013st} (STE) has been proposed as a method for learning in computation graphs involving a non-differentiable function $f$ of an input $x$, by replacing $f$ with the identity in the backward pass during computation of the gradients.

In practice, applying the STE for learning in \myMethod{} requires ensuring that the gradients flow through the sampled nodes, for example by treating the top-k sampled nodes at each layer as a mask which multiplies the messages from these nodes to \emph{all} their neighbors. Such a computation over all neighboring nodes defeats the purpose of sampling, increasing memory usage.

An alternative to keep memory usage low is to keep the gradients only for the top-k probabilities, though this results in a biased estimate of the gradient. We implemented two versions of this approach. In the first version (STE), we simply use the perturbed log probability (log probability plus the Gumbel noise) of the top-k nodes as additional weights for the classification GCN. In the second version, we normalized these weights to have a mean equal to one (STE$_\text{N})$. We present results in Table~\ref{tab:ste-comparison}. While in Cora the results are relatively similar to our implementation of \myMethod{}, as we experiment with larger graphs, there is a significant drop in performance when employing either variant of the STE.

\section{Proof of Theorem 1} \label{proof}
\subsection{Proof}
\begin{figure}[t]
    \centering
    \includegraphics[width=\textwidth, trim=21cm 10.5cm 24cm 17.5cm, clip]{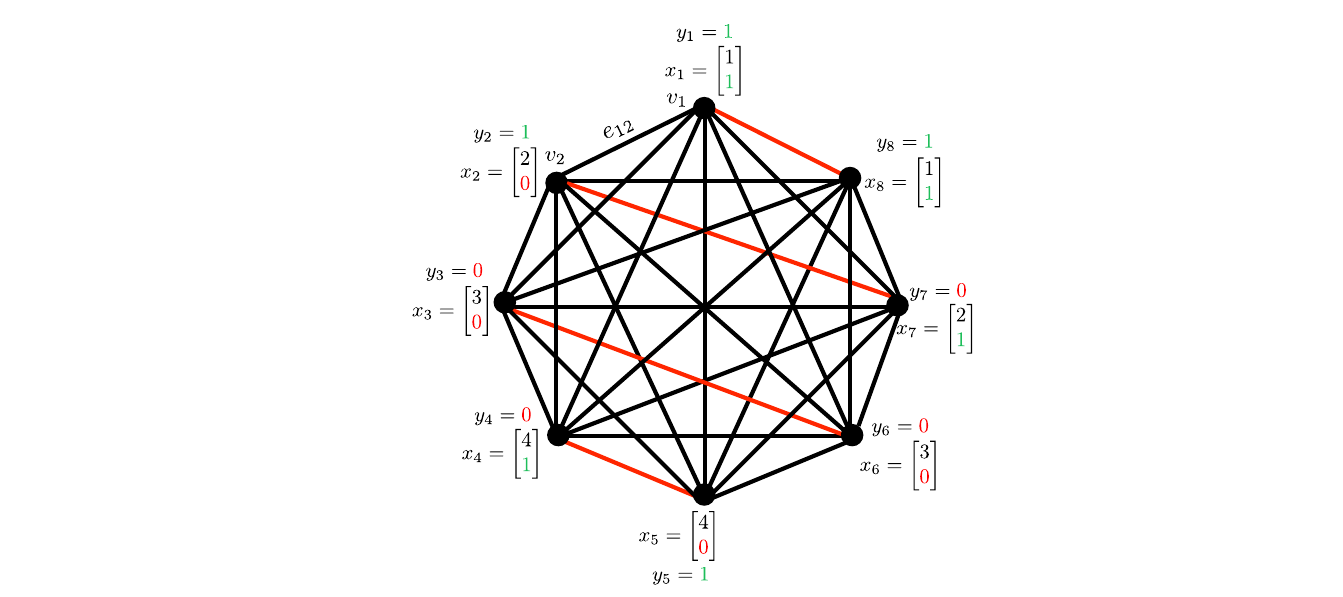}
    \caption{An example of a graph for Theorem 1 with eight nodes. Red edges belong to $\mathcal{E}_1$, features $x_i$ and labels $y_i$ are shown beside every node. For nodes $v_1$ and $v_2$ we show the edge $e_{12}$ as an example. As shown, the label of each node is the second feature of its neighbor, where a red edge connects them. The edge homophily ratio is $h=\frac{12}{28}=0.43$.}
    \label{fig:proof1}
\end{figure}

In this section, we prove Theorem \ref{theo1} in section \ref{theoretical analysis} as follows:
\begin{proof}
Let $\mathcal{G}=(\mathcal{V,E)}$ be an undirected fully connected simple graph with an even number of nodes N, $\mathcal{V}=\{1, \dots, N\}$, a set of edges $\mathcal{E}=\{e_{ij}\}_{i,j=1}^N$, where $e_{ij}$ denotes an edge between $v_i$ and $v_j$. 
Let $Y=\{y_i|i \in \{1,\dots,N\}, y_i \in \{0, 1\}\}$ be the set of node labels and let $X \in \mathbb{R}^{N \times 2}$ be the node features. 
Let $\mathcal{E}_1$ and $\mathcal{E}_2$ be one partition of $\mathcal{E}$, where $\mathcal{E}_1$ consists of $N/2$ edges with two conditions: 1) $e_{ij} \in \mathcal{E}_1$, if and only if the first feature of nodes $v_i$ and $v_j$ are equal, i.e. $e_{ij} \in \mathcal{E}_1 \Leftrightarrow x_{i1}=x_{j1}$, and 2) each node appears only at one edge in $\mathcal{E}_1$. 
The second condition implies that each value for the first feature appears only in the nodes connected by one edge in $\mathcal{E}_1$, i.e., for two distinct edges $e_{ij},e
_{kl} \in \mathcal{E}_1$, $x_{1i} \neq x_{1k}$. 

For each edge $e_{ij}$ in $\mathcal{E}_1$, sample $x_{i2} \sim\text{Bern}(0.5)$ on $\{0,1\}$. 
Then, set the node label $y_{j}$ to $x_{i2}$. That is, the label of each node is equal to the second feature of its corresponding neighbor in $\mathcal{E}_1$. 
% where $\exists \, \{f_k\}_{k=1}^{N/2} \in \mathbb{R}$ such that $\forall k, \; \left| \left\{ i \in \{1, \dots, N\} \mid x_{i1} = f_k \right\} \right| = 2$ and $x_{i2} \in \{0,1\}$, where $\mathbb{P}(x_{i2}=0)=0.5$. Let $y_i=x_{j2}$, where $x_{i1}=x_{j1}$. 
Since $\mathcal{G}$ is fully connected, the $L$-hop neighborhood for all nodes is $\mathcal{V}$.
% adjacency matrix $A=\mathbf{1}_{N \times N}$
Therefore, non-adaptive methods cannot prefer any node over another:  a GCN with a non-adaptive sampler can only achieve accuracy above chance by sampling $K$ nodes such that for each target node, the neighbor with the same first feature is sampled. 
The probability of sampling such a neighbor is $p=\frac{\binom{N-2}{K-1}}{\binom{N-1}{K}}=\frac{K}{N-1}$. The nominator indicates choosing the desired neighbor connected to the target node in $\mathcal{E}_1$ and then choosing the remaining $K-1$ nodes among the $N-2$ nodes. The denominator indicates choosing $K$ nodes among $N-1$ nodes (excluding the target node). 
Repeating the sampling for $L$ layers, eliminating $K$ nodes from the candidate neighbors for sampling for the next layer, will result in $p=\sum_{i=0}^{L-1}{\frac{K}{N-iK-1}}\leq\frac{LK}{N}$. 

However, there is a GCN with an adaptive sampler that classifies nodes with an accuracy of $100\%$. 
Specifically, such an adaptive sampler compares the first node feature of the neighbors with the target node $v_i$, and then always samples the unique $v_j$ where $e_{ij} \in \mathcal{E}_1$. 
Subsequently, it samples $K-1$ additional neighbors randomly. If the sampler perfectly samples the informative neighbors in $\mathcal{E}_1$, we assume there exists a GCN that can classify the target nodes correctly. We show a construction of such a GCN in section \ref{construction}.
Figure \ref{fig:proof1} shows an example of such a graph with eight nodes and a homophily ratio of $0.43$. 
\end{proof}

In the above proof, we show that there are graphs for which there is a performance gap between adaptive and non-adaptive samplers. Usually $LK \ll N$; therefore, the probability $p\leq\frac{LK}{N}$ for non-adaptive sampling methods to perform better than the chance level becomes very small. These graphs are heterophilous, with an expected homophily ratio equal to $0.5$. We calculate this expectation below:
\[\mathbb{E}(h_\mathcal{G})=\mathbb{E}\left(\frac{\binom{m}{2}+\binom{N-m}{2}}{\binom{N}{2}}\right), m \sim \text{Binomial}(N,0.5) \]

For that, we need $\mathbb{E}\left(\binom{m}{2}\right)=\frac{1}{2}\mathbb{E}\left(m(m-1)\right)$. 
Because $\mathbb{E}(m)=\frac{N}{2}$ and $Var(m)=\frac{N}{4}$, we have 
\[\mathbb{E}(m(m-1))=\mathbb{E}(m^2-m)=\mathbb{E}(m^2)-\mathbb{E}(m)=Var(m)+\left(\mathbb{E}(m)\right)^2-\mathbb{E}(m)\]

\[=\frac{N}{4}+\left(\frac{N}{2}\right)^2-\frac{N}{2}=\frac{N^2-N}{4},\]

so we have 
\[\mathbb{E}\left(\binom{m}{2}\right)=\frac{1}{2}\times\frac{N^2-N}{4}=\frac{N^2-N}{8}.\]

Similarly, $\mathbb{E}\left(\binom{N-m}{2}\right)=\frac{N^2-N}{8}$. Therefore, the expected homophily ratio is 
\[\mathbb{E}(h_\mathcal{G}) = 
 \frac{\frac{N^2-N}{8}+\frac{N^2-N}{8}}{\binom{N}{2}}=0.5.\]

\subsection{GCN Construction} \label{construction}
In this section, we construct a GCN that can compute the classification rule assumed in the proof of Theorem 1.
Let $\mathbf{x}_i\in\mathbb{R}^2$ be an input feature vector of a node $v_i\in\mathcal{V}$. We construct a GCN where the first layer computes a 4-dimensional vector for a target node $v_i$:
\begin{equation}
    \mathbf{h}_i^{(1)} = \tilde{\mathbf{W}}_1 \mathbf{x}_i + \frac{1}{\vert\mathcal{N}(v_i)\vert} \sum_{v_j\in\mathcal{N}(v_i)} \mathbf{W}_1\mathbf{x}_j.
\end{equation}
Here, $\tilde{\mathbf{W}}_1$ is the self-loop weight, and $\mathbf{W}_1$ is the weight for the neighbors of a target node. We define the following:
\begin{equation}
    \tilde{\mathbf{W}}_1 = 
\begin{bmatrix}
1 & 0 \\
0 & 1 \\
0 & 0 \\
0 & 0 
\end{bmatrix}, \quad
\mathbf{W}_1 = 
\begin{bmatrix}
0 & 0 \\
0 & 0 \\
1 & 0 \\
0 & 1 
\end{bmatrix}
\end{equation}
For a given target node, the output of this layer is a vector $\mathbf{h}_i^{(1)}\in\mathbb{R}^4$ where the first two values correspond to the target node's features, and the other two values are the average of the features of its neighbors.
The next layer implements the following function: if values $h_{i1}^{(1)}$ and $h_{i3}^{(1)}$ are the same, then the output is the second value of this vector, otherwise it is zero. We denote this function as $h_i^{(2)} = f(\mathbf{h}_i^{(1)})\in\mathbb{R}$, which can be implemented with a multi-layer perceptron due to its universal approximation properties~\citep{hornik1989universal}. An MLP can be implemented in a GCN by omitting the neighbors and using a self-loop weight matrix only.
The last layer of the GCN selects the average value in the neighborhood of a target node from the scalars computed by $f$, which can be computed using a simple GCN where the self-loop weight is set to zero: $h_i^{(3)} = \frac{1}{\vert\mathcal{N}(v_i)\vert}\sum_{v_j\in\mathcal{N}(v_i)} h_j^{(2)}$. This output corresponds to the classification rule required for Theorem 1.

\end{document}